\documentclass{article}[11pt]
\usepackage[backend=biber,style=ieee,firstinits=true]{biblatex}
\usepackage{fullpage}
\usepackage{algorithm}% http://ctan.org/pkg/algorithms
\usepackage{standalone}
\usepackage{amsmath,amssymb}
\usepackage{amsthm}
\usepackage[caption=false]{subfig}
\usepackage{pgfplots}
\pgfplotsset{compat=newest}
\pgfplotsset{every axis legend/.append style={%
cells={anchor=west}}
}
\usepgfplotslibrary{polar}
\usetikzlibrary{arrows}
\tikzset{>=stealth'}
\usepgfplotslibrary{groupplots}
\usepackage{listings}
\usepackage{tikz}
\usepgfplotslibrary{groupplots}
\newcommand{\ENCMIN}[1]{\ensuremath{ \left \{ #1 \right \}}}

\newtheorem{theorem}{Theorem}
\DeclareMathOperator*{\TV}{TV}

\DeclareMathOperator*{\minimize}{\text{minimize}}
\DeclareMathOperator*{\subject_to}{\text{subject~to}}

\DeclareMathOperator*{\argmin}{arg\,min}
\newcommand{\reals}{\mathbf{R}}
\newcommand{\name}{saturating splines }

\newcommand{\inputstyle}[1]{
\begin{@empty}

\pgfplotsset{ytick = {0}, xtick={0,1}, xmin=-0.1, xmax=1.1}

\input{#1}

\end{@empty}
}

\begin{document}
\title{Saturating Splines and Feature Selection}
\author{Nicholas Boyd
\and Trevor Hastie
\and Stephen Boyd
\and Benjamin Recht
\and Michael I.\ Jordan}
% \author{\name Nicholas Boyd \email nickboyd@berkeley.edu
% \AND \name Trevor Hastie \email hastie@stanford.edu
% \AND \name Stephen Boyd \email boyd@stanford.edu
% \AND \name Benjamin Recht \email brecht@berkeley.edu
% \AND \name Michael I.\ Jordan \email jordan@cs.berkeley.edu}
% \editor{}
% \begin{titlingpage}
    \maketitle

% \end{titlingpage}
\begin{abstract}{}
We extend the adaptive regression spline model by incorporating
\emph{saturation},
the natural requirement that a function extend as a constant outside a certain range.
We fit saturating splines to data using a convex optimization problem over a space of measures,
which we solve using an efficient algorithm based on the conditional gradient method.
Unlike many existing approaches, our algorithm solves the original infinite-dimensional (for splines of degree at least two) optimization problem
without pre-specified knot locations.
We then adapt our algorithm to fit generalized additive models with \name as coordinate functions
and show that the saturation requirement allows our model to simultaneously perform feature selection and
nonlinear function fitting.
Finally, we briefly sketch how the method can be extended to higher order
splines and to different requirements on the extension outside the data range.
\end{abstract}

\section{Introduction}

Splines ---  piecewise polynomials with continuity constraints --- are widely used to fit data~\cite[\S5.1]{ESL}.
One issue with piecewise polynomials is that they behave erratically beyond their boundary
knot points, and
(typically) grow without bound outside of that range~\cite[\S5.2]{ESL}.
This instability makes extrapolation dangerous; practitioners must take care to avoid querying
spline models near or outside of the range of the training data.

Smoothing spline algorithms~\cite{DeBoor, Wahba, Green} ameliorate  this problem by fitting \emph{natural splines}, which
reduce to a lower-degree polynomial beyond the boundary knots.
The most commonly used varieties of smoothing splines are cubic smoothing splines
(degree-three splines that reduce to linear outside the boundary  knots) and
linear smoothing splines, which extend as constant.  The saturating splines we propose
are closely related to linear smoothing splines.

Smoothing splines use an $\ell_2$ or quadratic notion of complexity,
and hence fit models with a predetermined and dense set of knot points~\cite[\S5.4]{ESL}.
\emph{Adaptive regression splines}~\cite{mammen}, on the other hand, use an $\ell_1$-type penalty,
which can result in a  sparse set of adaptively chosen knots.
However, adaptive regression splines do not reduce to lower degree outside of the range of
their largest knots, and hence may suffer from instability.

We propose fitting adaptive regression splines with explicit constraints on the degree of
the spline outside of a certain interval. We call such splines \emph{saturating splines}.
While the approach we take can be extended to fitting splines of arbitrary degree
with constraints on arbitrary derivatives, in this paper we focus on fitting
linear splines that are flat (constant) outside the data range; we mention the extension
to higher degree splines in~\S\ref{s_higher_degree}.
We show that saturating splines inherit the knot-selection property of
adaptive regression splines, while at the same time behave like natural splines near the boundaries of the data.

We also show a very important benefit of our approach in the context of fitting generalized
additive models~\cite{hastie1990generalized} with
saturating spline coordinate functions: the saturation constraint naturally results in variable selection.
Not only do we control the complexity of each coordinate function through knot selection,
but with the saturation condition, no knots on a variable means the variable is out of the model. This is not true for adaptive splines, since the linear term is unpenalized
and hence each variable would always be in the model.
The lack of feature selection can
 hurt interpretability and, in certain cases, generalization. The saturation constraint we propose precludes linear functions, and in concert with the adaptive spline $\ell_1$ penalty encourages coordinate functions to be identically zero.
 As a result, generalized additive models fit with saturating spline component functions often depend on only a few input features.

Like smoothing splines and adaptive regression splines, saturating splines arise as solutions
to certain natural functional regression problems.
We solve the saturating spline fitting problem by reformulating it as a convex optimization problem over a space of measures, roughly speaking, the second derivative of
the fitted function. To the best of our knowledge, this approach is novel.
We then apply a variant of the classical conditional
gradient method~\cite{jaggi,adcg} to this problem.
At each iteration of our algorithm, an atomic measure is produced; moreover, we can uniformly bound the number of
atoms, which corresponds to the number of knot points in the spline function.
(While we manipulate atomic measures, we solve the problem over the space of all measures with
finite total variation.)
In contrast to standard coordinate descent methods,
in each iteration of the conditional gradient method the weights of {\em two} knot points are
adjusted.
In the fully corrective step, we solve a finite-dimensional convex optimization problem
with $\ell_1$ and simple linear constraints.
Numerical experiments show that the method is extremely effective in practice.

Our optimization method can exploit warm starts, i.e., it can use an initial guess
for the fitted function.
This allows us to compute an entire regularization path efficiently, at a cost typically just a small
multiple of the effort to solve the problem for one value of the regularization parameter.
Because our algorithm is based on the conditional gradient method, we can use the framework
of~\cite{path} to compute a provably $\epsilon$-suboptimal approximate regularization path.
When fitting generalized additive models, the regularization path has attractive features:
at critical values of the regularization parameter, new regressors are brought into (or, occasionally,
out of) the model, or new knot points are added to (or deleted from) one of the existing
coordinate functions.  Thus our approach combines feature selection and knot point selection.

\subsection{Outline}
In \S\ref{s_uni} we introduce a univariate function fitting problem,
inspired by the adaptive spline estimation problem of~\cite{mammen},
that includes the additional requirement that the fitted function saturate.
In \S\ref{s_splines} we make the connection between our function estimation problem and
standard adaptive splines, and pose the saturating spline fitting problem as a
convex optimization problem over measures.
In \S\ref{s_cg} we modify the classical conditional gradient method to solve this optimization problem.
In \S\ref{s_gam} we extend the optimization problem and algorithm to
fit generalized additive models to multivariate data.
We illustrate the effectiveness of the method with several examples in \S\ref{s_examples}.
We discuss generalizations to higher-degree splines in \S\ref{s_higher_degree}.
Finally, we discuss potential extensions and variations in \S\ref{s_extensions}.
The appendix includes implementation details and proofs.

\section{Univariate function fitting}\label{s_uni}
We wish to fit a continuous bounded function $f : \reals \rightarrow \reals$ from
data $(x_i,y_i) \in \reals \times \mathcal Y$, $i=1, \ldots, n$,
$x_i \in [0,1]$.
To do this we will choose $f$ to minimize a data mismatch or loss
function subject to a constraint that encourages regularity in $f$,
and an additional constraint, saturation, that we describe below.

The loss is given by
\[
L(f) = \sum_{i=1}^n \ell(f(x_i), y_i),
\]
where $\ell: \reals \times \mathcal Y \to \reals$ is nonnegative,
twice differentiable, and strictly convex in its first argument.
Typical loss functions include $\ell(z,w) = (z-w)^2/2$ (standard regression, $\mathcal{Y} = \reals$), or
$\ell(z,w) = \log (1+ \exp -(zw))$ (logistic regression, with $\mathcal Y = \{-1,1\}$).
The loss $L$ is a convex functional of the function $f$ that
only depends on the values of $f$ at the data points $x_i$.
The smaller the loss, the better $f$ fits the given data.

We constrain the function $f$ to be simple by limiting the
value of a nonnegative regularization functional $R$.
In this paper, we take $R$ to be the total variation of the
derivative of $f$,
\[
R(f) = \TV(f'),
\]
a convex functional of $f$.
For a twice-differentiable function $f$, recall that
\begin{equation}
	\label{tv_l1}
\TV(f') = \int | f''(x) | \; dx,
\end{equation}
i.e., the regularization is the $\ell_1$ norm of the
second derivative.
(As we review in the following section, the modern definition of total variation extends this equality to
nondifferentiable functions.)
The total variation limit we impose on $f$ is $R(f) \leq \tau$,
where $\tau$ is a parameter that we use to trade off model fit and model
regularity.
This regularization constraint implicitly constrains $f$ to be
differentiable almost everywhere, with its derivative having finite total variation.

Our model $f$ will be subject to one more constraint, that it
saturates (outside the interval $[0,1]$), which means that it is
a (possibly different) constant on the two intervals outside $[0,1]$:
$f(x)=f(0)$ for $x \leq 0$, and $f(x)=f(1)$ for $x \geq 1$.
In other words, $f$ extends as a constant outside the nominal data range of $[0,1]$.
In terms of the derivative, this is equivalent to the requirement that $f'$ exists and is zero outside $[0,1]$.

The fitting problem is then
\begin{equation}
\label{problem}
\begin{array}{ll}
\minimize & L(f) \\
\subject_to & R (f) \le \tau, \\
& f'(x) = 0 \mbox{ for } x \not\in [0,1],
\end{array}
\end{equation}
where $\tau \geq 0$ is the regularization parameter.
The variable to be determined is the function $f$,
which is in the vector space of continuous functions
with derivatives of finite total variation.
This fitting problem is an infinite-dimensional convex optimization problem.

In applications the problem~\eqref{problem} is solved for a range of values of
$\tau$, which yields the regularization path.
The final model is selected using a hold-out set or cross-validation.
For $\tau=0$, $f$ must be constant and the problem~\eqref{problem}
reduces to fitting the best constant to the
data. As $\tau$ increases, $f$ is less constrained, and our fitted
model becomes more complex; eventually we expect overfitting.
For example, in the case of regression, with a loss function that satisfies
$\ell(u,u)=0$ and data with distinct $x_i$, the fitting function is the
piecewise-linear function that interpolates the data, for large enough $\tau$.

\section{Splines and functions of bounded variation}\label{s_splines}

In this section we explore the connection between our fitting problem and
degree-one splines, i.e., piecewise-linear continuous functions,
which have the form
\begin{equation}\label{e-fo-spline}
f(x) = c + \sum_{i=1}^K w_i (x - t_i)_+,
\end{equation}
where $(z)_+ = \max\{z,0\}$.  We assume that the $t_i$ are distinct, and
refer to them as knot points or simply knots.  The scalars $w_i$ are the weights,
and $c$ is the offset.
We refer to the function $x \mapsto (x-t_i)_+$ as a hinge function, so a
degree-one spline is a finite linear combination of hinge functions, plus
a constant.

\subsection{Functions of bounded variation}
A right-continuous function $h : [0,1] \rightarrow \reals$
is of bounded variation if and only if there exists a
signed measure $\mu$ on $[0,1]$ with
\begin{equation}
\label{total_variation}
h(z) = \int 1(y \le z)  \; d\mu(y),
\end{equation}
where $1(y \leq z)=1$ for $y\leq z$ and $0$ otherwise.
The measure $\mu$ is unique; we can think of it as the derivative of $h$.
That is,~\eqref{total_variation} is essentially the second fundamental
theorem of calculus with $h'$ replaced by $\mu$.

We also have $\TV(h) = \int d |\mu|$.  (This is called the total variation
of the measure $\mu$.)
We will denote this using the notation $\| \mu \|_1$,
to emphasize the similarity with the finite-dimensional case, or the case
when $h$ is differentiable: $\TV(h) = \|h'\|_1$.
When the measure $\mu$ is atomic, the function $h$ is piecewise
constant with jumps at the points in the support of $\mu$.

\subsection{Splines and derivatives with bounded variation}
Now suppose that $f : [0,1] \rightarrow \reals$ has a right-continuous {\em derivative\/} of bounded variation.
From~\eqref{total_variation}, with $h = f'$, and the fundamental theorem of calculus,
we have
\begin{align}
\label{spline}
f(x) &= f(0)+ \int_0^x f'(z) \; dz = f(0)+ \int_0^x \int 1(y \le z) \; d \mu(y)\;  dz \\
&= f(0)+ \int \int_0^x 1(y \le z) \; dz \; d \mu(y) \\
&= f(0) + \int  (x - y)_+\;  d \mu(y).
\end{align}
This shows that any such function is a (possibly infinite) linear combination of hinge functions,
plus a constant (i.e., $f(0)$).
In this case, the measure $\mu$ can be thought of as the \emph{second} derivative of $f$.

When $\mu$ is atomic and supported on a finite set, that is,
\[
\mu = \sum_{i=1}^K w_i \delta_{t_i},
\]
$f$ is a degree-one spline of the form~\eqref{e-fo-spline},
with $c = f(0)$.
So degree-one splines correspond exactly to the case where the measure $\mu$
(roughly, the second derivative) has finite support.

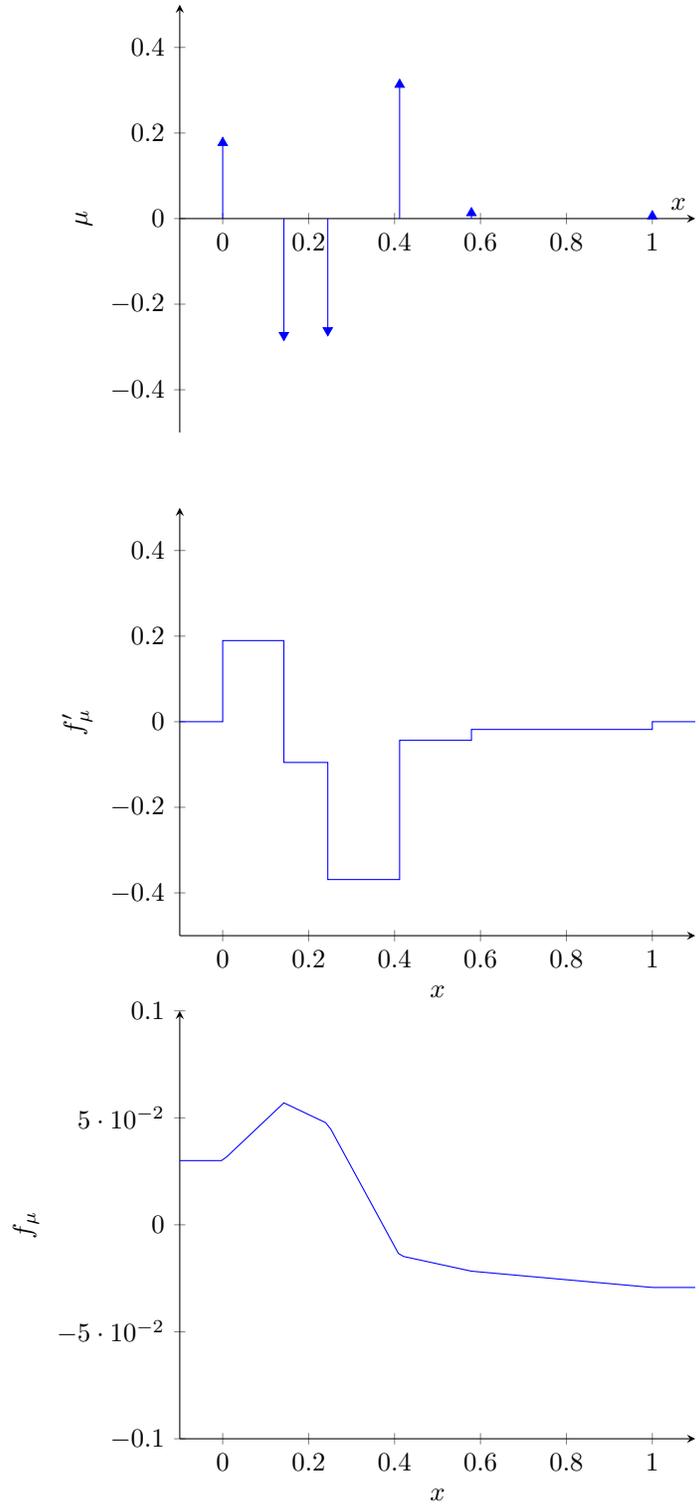
\begin{figure}
\centering
\begin{tikzpicture}
\begin{groupplot}[group style={vertical sep = 1cm, group size= 1 by 3}]
	\nextgroupplot[
	    axis lines = left,
	    xlabel = $x$,
			ylabel = $\mu$,
			xmax = 1.1,
			xmin = -0.1,
			ymax = 0.5,
			ymin = -0.5,
			axis x line=center,
	]
	\addplot +[mark=triangle*,
	   mark options={scale=1},
	   scatter,ycomb,
	   visualization depends on={(y-abs(y))/abs(y-abs(y)) \as \sign},
	   scatter/@pre marker code/.code={\scope[rotate=180*\sign,yshift=-2pt]}
	   ] coordinates {(0.0, 0.1895222749219244)  (0.1424148606811145, -0.28473011635102846)  (0.24458204334365322, -0.2736687705893021)  (0.411764705882353, 0.32539749281795594)  (0.5789473684210527, 0.025390182656613774)  (1.0, 0.018088936543836594)};

\nextgroupplot[
    axis lines = left,
    xlabel = $x$,
		ylabel = $f_\mu'$,
		xmax = 1.1,
		xmin = -0.1,
		ymax = 0.5,
		ymin = -0.5,
]
\addplot [
    color=blue,
]
coordinates {(-0.1,0)  (0.0,0.0) (0.0,0.1895222749219244)  (0.1424148606811145,0.1895222749219244) (0.1424148606811145,-0.09520784142910407)  (0.24458204334365322,-0.09520784142910407) (0.24458204334365322,-0.36887661201840616)  (0.411764705882353,-0.36887661201840616) (0.411764705882353,-0.04347911920045022)  (0.5789473684210527,-0.04347911920045022) (0.5789473684210527,-0.018088936543836448)  (1.0,-0.018088936543836448) (1.0,1.457167719820518e-16)  (1.1,0.0)};
\nextgroupplot[
	     axis lines = left,
	     xlabel = $x$,
	 		ylabel = $f_\mu$,
	 		xmax = 1.1,
	 		xmin = -0.1,
	 		ymax = 0.1,
	 		ymin = -0.1,
	 ]
	 \addplot [
	     domain=-0.1:1.1,
	     samples=100,
	     color=blue,
	 ]
	 {0.03 + 0.1895222749219244*max(x-0.0,0) + -0.28473011635102846*max(x-0.1424148606811145,0) + -0.2736687705893021*max(x-0.24458204334365322,0) + 0.32539749281795594*max(x-0.411764705882353,0) + 0.025390182656613774*max(x-0.5789473684210527,0) + 0.018088936543836594*max(x-1.0,0)};

\end{groupplot}
\end{tikzpicture}
\caption{$f_\mu$ and $ f_\mu'$ generated by the atomic measure $\mu$ ($f_\mu''$).
The regularization functional, $\TV(f_\mu')$, is the sum of the absolute values of the spikes in $\mu$.
Note that the (signed) sum of the spikes in $\mu$ is zero: that is, $\int d\mu = 0$, which implies that $f_\mu$ saturates.}
\label{f_fmu}
\end{figure}

We introduce the notation
\begin{equation}
\label{fmu}
f_\mu(x) = \int_0^x \int 1(t \le z)  \;d\mu(t) \; dz = \int {(x - t)}_+\;d\mu(t)
\end{equation}
to denote the function
derived from the measure $\mu$.  It is, roughly speaking, the
double integral of the measure $\mu$, or the (potentially infinite) linear combination of
hinge functions associated with the measure $\mu$.
The mapping from $\mu$ to $f_\mu$ is linear, and we have $\TV(f'_\mu) = \|\mu\|_1$.
A simple example of $f_\mu$, its first derivative $f_\mu'$, and its (atomic measure)
second derivative $\mu$ is shown in Figure~\ref{f_fmu}.

\clearpage
\subsection{Fitting splines by optimizing over measures}
\label{ss_overmeasures}
Identifying $f = c + f_\mu$,
we can solve the fitting problem~\eqref{problem} by minimizing
over the bounded measure $\mu$ on $[0,1]$, and the constant $c$.
The measure $\mu$ is the second derivative of $f$, and the constant $c$
corresponds to $f(0)$.
The total variation regularization constraint $\TV(f') \leq \tau$
corresponds to $\|\mu\|_1 \leq \tau$.
The saturation condition holds by construction for $x<0$; to ensure that
$f'(x)=0$ for $x>1$, we need
\[
f'(1) = f'(0)+ \int_0^1 d \mu = 0.
\]
In other words, saturation of $f$ corresponds to $\mu$ having total (net) mass zero.
Thus~\eqref{problem} can be rephrased as
\begin{equation}
\label{problem-nu}
\begin{array}{ll}
\minimize & L(E_x \mu +  c) \\
\subject_to & \| \mu \|_1\le \tau, \\
& \int  d \mu = 0
\end{array}
\end{equation}
over the bounded measure $\mu$ on $[0,1]$, and $c \in \reals$.
Note the slight abuse of notation here: we now (and for the remainder of the paper) consider $L$ as a functional on $\reals^n$.
In the above, $E_x$ is the linear operator that maps $\mu$ to the vector $(f_\mu(x_1), \ldots, f_\mu(x_n))$, given by~\eqref{fmu}.
$E_x$ is clearly linear, as it is the integral of the function $\psi : \reals \rightarrow \reals^n$: \[\psi(t) = ((x_1 - t)_+, \ldots, (x_n -t)_+)\] against $\mu$.
We will apply the conditional gradient method directly to this problem.

To gain intuition about the optimization problem~\eqref{problem-nu}, we can consider it as a infinite-dimensional analogue
of the standard lasso~\cite{lasso}. The lasso is the solution to the optimization problem
\begin{equation}
\label{problem-lasso}
\begin{array}{ll}
\minimize & \frac{1}{2}\| Aw -y \|_2^2 \\
\subject_to & \| w \|_1\le \tau.
\end{array}
\end{equation}
Here $w$ is a vector in $\reals^d$, and $A \in \reals^{(n,d)}$ is a matrix.
Ignoring the constant term $c$, we see that~\eqref{problem-nu} looks very similar to~\eqref{problem-lasso}, where $E_x$ plays the role of $A$; indeed,
$E_x$ is essentially a matrix with $n$ rows and infinitely many columns.
Our intuition from the lasso
suggests that there should be solutions of~\eqref{problem-nu} that are
sparse, which here means that $\mu$ is atomic.  In terms of $f_\mu$,
sparsity means there are solutions of the original functional fitting problem
\eqref{problem} that are degree-one splines.
This is indeed the case.
Theorem~\ref{theorem_1}
shows that there is a solution of
\eqref{problem-nu} with $\mu$ atomic, supported on no more than $n+2$
points; in other words, $f_\mu$ is a degree-one spline with $K \leq n+2$.
Moreover, in practice the solution of~\eqref{problem-nu} will exhibit
selection, that is, it will be supported on far fewer than $n+2$ points.

\begin{theorem}{}
\label{theorem_1}
 Fix $x_1, \ldots, x_n \in [0,1]$ and $f : \reals \rightarrow \reals$ with $f'$ (right-continuous) of bounded total variation, and $f$ constant outside of $[0,1]$.
 Then there exists a degree-one saturating spline $\hat{f}$ (with an most $n+2$ knots) that matches $f$ on $x_i$ with $\TV(\hat{f}') \le \TV(f')$.
 \end{theorem}

For the remainder of the paper we will ignore the constant term $c$.
It is not difficult to adapt the algorithms we present to handle the constant
term, but doing so does add some notational complexity.  It's also possible to minimize out
$c$, as it does not affect the regularization term;
the resulting problem is still convex in $w$.

\section{The conditional gradient method for fitting splines}\label{s_cg}
In this section we outline our algorithm for solving~\eqref{problem-nu} (and therefore also~\eqref{problem}).
To that end, we briefly review the classical conditional gradient method~\cite{jaggi} and the
measure-theoretic version proposed in~\cite{adcg}.

The optimization problem we need to solve,~\eqref{problem-nu}, (without the constant term $c$) is
\begin{equation}
\label{infinite}
\begin{array}{ll}
\minimize &  L(E_x \mu) \\
\subject_to & \int d \mu = 0, \\
& \|\mu \|_1 \le \tau.
\end{array}
\end{equation}
As noted in the last section,~\eqref{infinite} is a convex optimization problem over a space of measures.
We closely follow the approach taken in~\cite{adcg} and apply the conditional gradient method to this problem directly.

The main benefit of this approach is that we can restrict our attention to \emph{atomic} measures,
i.e., $\mu$ of the form
\[ \mu = \sum_{j =1}^K w_j \delta_{t_j}.\]
Measures of this form are easily representable in a computer, by simply storing
a list of $(w_j, t_j)$ pairs.
Theorem~\ref{theorem_1} ensures that the number of knots we need to store is absolutely
bounded, i.e., that our algorithm runs in bounded memory.
While we manipulate atomic measures, we solve the problem~\eqref{infinite} over all bounded
measures.

One thing to note about finitely-supported atomic measures is that we can easily optimize over the
weights $w_j$ with the knot locations $t_j$ fixed, since this corresponds to a
finite-dimensional convex optimization problem amenable to any standard
algorithm. Our algorithm makes use of this fact, and alternates between adding
pairs of knots and optimizing over the weights $w$ at each iteration.
In this latter step knots can be (and indeed eventually must be) removed.
In an additional and optional step the knot points can be moved
continuously within $[0,1]$, or to neighboring data points.
This step is not needed for theoretical convergence but can improve convergence and the sparsity of the final solution in
practice.

\subsection{The conditional gradient method}

The conditional gradient method (CGM) solves constrained convex optimization problems of the form
\begin{equation}
\label{cgm_problem}
\begin{array}{ll}
\minimize & f(x) \\
\subject_to & x \in \mathcal{C},
\end{array}
\end{equation}
with variable $x \in \reals^d$. In the above, it is always assumed that the (convex) function $f$ is differentiable.
At each iteration of the CGM we form the standard linear approximation to the function $f$ at the current iterate $x_m$:
\[\hat{f}(x; x_m) =  f(x_m) + f'(x - x_m; x_m).\]
Here $f'(d;x)$ is the directional derivative of the function $f$ at $x$ in the direction $d$, defined by
\[ f'(d;x) = \lim_{t\searrow 0} \frac{f(x + td) - f(x)}{t}.\]
Our use of the directional derivative here may seem surprising: for differentiable functions on $\reals^d$, $f'(d;x)$ is always equal to $\langle \nabla f(x), d \rangle.$
The direct applicability of directional derivatives to convex functionals of
measures motivates us to prefer the directional derivative.

Convexity of $f$ implies that $\hat{f}$ is a \emph{lower} bound on $f$, that is:
\begin{equation}
\label{lower_bound}
\hat{f}(x; x_m) \le f(x).
\end{equation}

In the next step of the CGM, we minimize this first-order approximation over the feasible set
$\mathcal{C}$:
\[ s_m \in \argmin_{s \in \mathcal C} \hat{f}(s;x_m)  = \argmin_{s \in \mathcal C} f'(s;x_m). \]
The point $s_m$ is called the conditional gradient of $f$. Note that $s_m$ provides a lower bound on $f(x_\star)$:
\[ \hat{f}(s_m; x_m) \le f(x_\star).\]
In particular, we can bound the sub-optimality of the point $x_m$:
\begin{equation} \label{bound} f(x_m) - f(x_\star) \le - f'(s_m-x_m; x_m). \end{equation}
One can show (as in ~\cite{jaggi}) that this bound decreases to zero, which means that it can be used as a (non-heuristic)
termination criterion.
After determining $s_m$, there are several options for updating $x_m$.
In this paper, we will use the fully-corrective variant of the CGM,
which chooses $x_{m+1}$ to minimize $f$ over the convex hull of $\{s_1, s_2, \ldots, s_m\}$.
Note that this last step may become computationally intensive as $k$ grows, and indeed limits the applicability of the conditional gradient method to problems where this step is computationally feasible.
One option is to remove previous conditional gradients as soon as they are not selected in the minimization step.
Caratheodory's theorem ensures us that the set of previous conditional gradients we need to track is then bounded by $d+1$.
In practice, however, the algorithm is usually terminated well before $d+1$ iterations.

\begin{minipage}{.8\textwidth}
\begin{algorithm}[H]
\caption{Fully-corrective conditional gradient method}
\smallskip
{\bf For} $m = 1,\ldots$
%{\bf while} {$\hat{f}(s_k,x_k) > \tol$}
\begin{enumerate}
\item Linearize: $\hat{f}(s;x_m)  \leftarrow f(x_{m}) + f'(s - x_{m};x_{m})$. \label{linearize}
\item Minimize: $s_m \in \arg\min_{s\in \mathcal{C}} \hat{f}(s;x_m)$.  \label{lmo}
\item Update: $x_m \in \arg\min_{x \in \textrm{conv}(s_1, \ldots, s_m)} f(s).$
\end{enumerate}
\end{algorithm}
\end{minipage}

\begin{figure}
\begin{center}
\begin{tikzpicture}
	\begin{axis}[	xmin = -0.5, xmax = 2, ymin = -0.5 ,ymax = 1.5,legend pos=north west]
		\addplot[samples=100] {x^2};
		\addlegendentry{$f(x)=x^2$}
		\addplot[style=dotted] {x-1/4};
		\addlegendentry{$\hat{f}(\cdot;\frac{1}{2})$}
		\addplot[style=dashed] coordinates {(0.5,-1) (0.5,3)};
		\addlegendentry{$x_m = \frac{1}{2}$}
		\addplot[] coordinates {(-0.25,-1) (-0.25,3)};
		\addplot[] coordinates {(1.25,-1) (1.25,3)};

	\end{axis}
\end{tikzpicture}
\end{center}
\caption{An illustration of a single iteration of the conditional gradient method on the function $f(x) = x^2$ at the point $\frac{1}{2}$.
The set $\mathcal{C}$ is the interval $[-0.25,1.25]$, indicated by the solid vertical lines.
The first order approximation $\hat{f}(\cdot; \frac{1}{2})$ is plotted as the dotted line tangential to $f(x)$ at $\frac{1}{2}$.
The conditional gradient $s_m$ is the point $-0.25.$}
\end{figure}
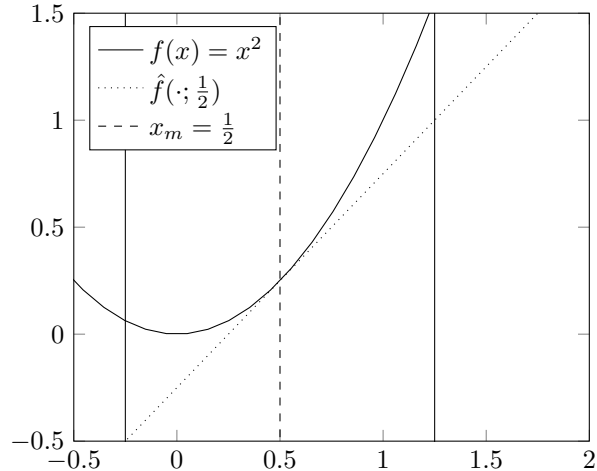

\subsection{Conditional gradient for measures}

In this subsection, we apply the conditional gradient method to the infinite-dimensional
problem~\eqref{infinite}, which we repeat here:
\begin{equation}
\begin{array}{ll}
\minimize &  L(E_x \mu) \\
\subject_to & \int d \mu = 0, \\
& \|\mu \|_1 \le \tau.
\end{array}
\end{equation}

First we'll show that the conditional gradient, i.e., the measure $s_m$, can be chosen to be supported on exactly two points, and is computable in time linear in $n$.
The directional derivative of the objective function in the direction of the measure $s$ at the point $\mu$ is given by
\begin{align*}
&\lim_{t \searrow 0} \frac{ L(E_x (\mu + ts))  - L(E_x \mu)}{t} \\
		&= \lim_{t \searrow 0} \frac{ L(E_x \mu + t E_x s)  - L(E_x \mu)}{t} \\
		&= L'(E_x s; E_x \mu) \\
		&= \langle \nabla L(E_x \mu) , E_x s \rangle_{\reals^n}.
		% &= \int \langle \nabla L(E_x \mu), E_x \delta_t \rangle \; ds(t).
\end{align*}

We can then interchange the inner-product in $\langle \nabla L(E_x \mu) , E_x s \rangle$
with the integral in $E_x s = \int \psi(t)\;ds(t)$:
\begin{equation}
\langle \nabla L(E_x \mu) , E_x s \rangle = \int \langle \nabla L(E_x \mu), \psi(t) \rangle \; ds(t).
\end{equation}

Let $g = \nabla L(E_x \mu) \in \reals^n$. Note that in the case $\ell(x,y) = \frac{(x-y)^2}{2}$,
$g$ is simply the residual $E_x \mu - y$ and $\langle g, \psi(t) \rangle$ is the correlation
between the residual and a single hinge function located at $t$.
A conditional gradient is any solution to the following optimization problem
\begin{equation}
\label{cg}
\begin{array}{ll}
\minimize & \int \langle g, \psi(t) \rangle \; ds(t) \\
\subject_to & \int ds = 0,\\
		& \| s \|_1 \le \tau.
\end{array}
\end{equation}

Without the integral constraint, we would expect there to be a solution to~\eqref{cg} that is a single point-mass:
the objective function is the integral of a scalar-valued function against a bounded measure.
We'll show that there is always a solution to~\eqref{cg} that is supported on exactly two points.
Furthermore, we'll show that those two points can be computed in time linear in $n$.

First we'll construct a particular feasible point for~\eqref{cg} and then we'll show that
it achieves the optimal value.
Let \[t_+ \in \argmin_t   \langle g , \psi(t) \rangle,\;\ t_- \in \argmin_t   -\langle g , \psi(t) \rangle. \]
Define \[s_\star = \frac{\tau}{2} \delta_{t_+} - \frac{\tau}{2} \delta_{t_-}.\]
The objective value achieved by $s_\star$ is \[o_\star = \frac{\tau}{2}\left( \langle g , \psi(t_+) \rangle -\langle g , \psi(t_-) \rangle \right).\]
We'll show that either any measure $s$ that is feasible for~\eqref{cg} has objective value bounded below by $o_\star$ or $\mu$ is optimal for~\eqref{infinite}.
Let $s$ be any feasible measure for~\eqref{cg}. Decompose $s$ into the difference of two mutually singular non-negative measures:
$s = s_+ - s_-$. Then as $s$ is feasible we have $\|s_+\|_1 = \|s_-\|_1 \le \frac{\tau}{2}.$
The objective value achieved by $s$ can be bounded below as follows
\begin{align*}
\int \langle g, \psi(t) \rangle ds(t) &= \int \langle g, \psi(t) \rangle ds_+(t) + \int -\langle g, \psi(t) \rangle ds_-(t) \\
&\ge \|s_+\|_1 \left(\min_t  \langle g , \psi(t) \rangle\right) + \|s_-\|_1\left(\min_t  -\langle g , \psi(t) \rangle\right) \\
&\ge \|s_+\|_1 \left(\min_t  \langle g , \psi(t) \rangle + \min_t   -\langle g , \psi(t) \rangle \right).
\end{align*}
Suppose $\left(\min_t  \langle g , \psi(t) \rangle + \min_t   -\langle g , \psi(t) \rangle \right) \ge 0$.
Then the argument above implies $s_\star = 0$ is a conditional gradient for~\eqref{infinite}, and thus~\eqref{bound} implies $\mu$ is optimal.
Otherwise we have \[ \left(\min_t  \langle g , \psi(t) \rangle + \min_t   -\langle g , \psi(t) \rangle \right) < 0,\]
which implies
\[ \|s_+\|_1 \left(\min_t  \langle g , \psi(t) \rangle + \min_t   -\langle g , \psi(t) \rangle \right) \ge \frac{\tau}{2} \left(\min_t  \langle g , \psi(t) \rangle + \min_t   -\langle g , \psi(t) \rangle \right)  = o_\star.\]
This proves the assertion.

Note that finding $t_-$ and $t_+$ involves two \emph{separate} optimization problems over $[0,1]$ instead of one over $[0,1]\times[0,1]$.
These problems are readily solved by gridding, though in this case they can be solved exactly in time linear in $n$ if we have access to a sorted vector of the data points $x_i$.
To see this, we expand the objective function for $t_+$ above,
\[ t_+ = \argmin_{0 \le t \le 1}  \sum_{i=1}^n g_i (x_i - t)_+  = \argmin_t \sum_{i : x_i \ge t} g_i (x_i - t).\]
If $x_i$ are sorted, we can compute the minimizer between each pair of consecutive data points exactly,
since this is simply computing the minimizer of a linear functional over an interval. Thus
in a single pass over the data we can compute the global minimizer exactly.

Immediately after computing $t_-$ and $t_+$ we can use~\eqref{bound} to bound the suboptimality of $\mu$ by
\[L(E_x \mu) - L(E_x \mu_\star) \le - \int \langle g, \psi(t) \rangle d(s_\star-\mu)(t).\]

With this choice of conditional gradient,
the fully-corrective step is a finite-dimensional convex problem.
Fixing the knot locations encountered as conditional gradients so far, $t_1, \ldots, t_{2k}$,
we can do at least as well as the fully-corrective algorithm by solving the following optimization
problem:
\begin{equation}
\label{fix_supp}
\begin{array}{ll}
\minimize &  L(E_x \mu) \\
\subject_to & \int d \mu = 0, \\
& \|\mu \|_1 \le \tau,\\
& \textrm{supp}(\mu) \subset  \{ t_1, \ldots, t_{2k}\}.
\end{array}
\end{equation}
This is equivalent to the following optimization problem in $\reals^{2k}$:
\begin{equation}
\label{finite_sup}
\begin{array}{ll}
\minimize &  L(\sum_j w_j E_x \delta_{t_j}) \\
\subject_to & 1^Tw = 0, \\
& \|w \|_1 \le \tau.
\end{array}
\end{equation}
We can solve this using any of a number of existing algorithms~\cite{admm, spgl1}.
In our implementation we use the conditional gradient method with line-search for simplicity.

By warm starting with an increasing sequence of $\tau$'s, we can efficiently compute
 an approximate regularization path. Indeed we can even provide a provably $\epsilon$-suboptimal
 path using the approach of~\cite{path}.

\subsection{Convergence}
As in the case of ADCG~\cite{adcg} convergence follows immediately from the
conditional gradient method proof in general Banach spaces~\cite{DUNN,ZAMM,jaggi}.
The convergence of the conditional gradient method
depends on a curvature parameter $C_f$. $C_f$ is a constant such that the following inequality is satisfied for all $x, s \in \mathcal{S}$ and $\eta \in (0,1)$:
\[
\label{curvature}
f(x + \eta(s-x)) \le f(x) + \eta f'(s-x; x) + \frac{C_f}{2}\eta^2.
\]
For our purposes $f : \mathbf{R}^n \rightarrow \mathbf{R}$ is simply $L$ and
$\mathcal{S} = \{E_x \mu : \|\mu\|_1 \le \tau, \int d\mu = 0\}.$
A simple sufficient condition for $C_f$ to be finite is that $\ell$ is differentiable with Lipschitz gradient.
If $C_f$ is finite, the conditional gradient method converges (in terms of function value) at a rate of at least $1/m$ where $m$ is the iteration counter.

\section{Generalized additive models}\label{s_gam}
One natural application of univariate splines is fitting generalized additive models~\cite{hastie1990generalized} to multivariate data: $(x_i,y_i) \in \reals^D \times \mathcal Y$, $i=1, \ldots, n$.
That is, fitting a function of the form
\[ f(x) = \sum_{d=1}^D f_d(x[d]) \]
where each $f_d$ is a simple function from $\reals$ to $\reals$ (here $x[d]$ is the $d$-th coordinate of the vector $x$).
We can mimic our approach in the scalar case with the following optimization problem:
\begin{equation}
\label{gam_problem}
\begin{array}{ll}
\minimize & L(f) \\
\subject_to & \sum_d R (f_d) \le \tau, \\
& f_d'(x) = 0~\forall x \not\in [0,1], d.
\end{array}
\end{equation}

Here $R$ is the same regularizer used in the scalar case, namely \[ R(g) = \TV(g') \simeq \| g'' \|_1.\]
As in the scalar case, one can show that there is always an optimal $f$ with each coordinate function $f_d$ a degree-one saturating spline.

This allows us to rephrase~\eqref{gam_problem} as an optimization problem over measures.
The only change from the scalar case is that the measure is over the set $\{1, \ldots, D\} \times [0,1]$ --- each knot is now attached to a particular coordinate.
In other words, we search for a function of the following form:
\[ f_\mu (x) = \int {(x[d] - t)}_+ d\mu(d,t).\]

We again have equality between the $\ell_1$ norm of $\mu$ and the regularization term:
\[ \sum_d R({(f_\mu)}_d)  =  \| \mu \|_1 .\]

The analogue of~\eqref{infinite} is then
\begin{equation}
\label{gam}
\begin{array}{ll}
\minimize &  L(E_x \mu) \\
\subject_to & \int 1(d = \hat{d}) \;d\mu(d,t) = 0,\;\forall \hat{d} \\
& \|\mu \|_1 \le \tau.
\end{array}
\end{equation}
The conditional gradient algorithm from the scalar case generalizes immediately to fitting generalized additive models --- the only difference is that we now need to find a pair of knots for the same coordinate.
This involves solving $d$ pairs of nonconvex optimization problems over $[0,1]$ --- again this can be done by gridding or by sorting the training data.

Saturating splines gain an additional advantage over standard adaptive splines when fitting generalized additive models.
The addition of the saturation constraint (that $f_d$ be constant outside of $[0,1]$) naturally leads to variable selection when fitting generalized additive models.
What we mean by variable selection is that the functions $f_d$ are often exactly $0$.
This is because the saturation constraint means that linear coordinate functions no longer escape the regularization (indeed, they are impossible).
This is very different from the standard adaptive spline setup without the saturation constraint.
In that case, linear functions, i.e. $f_d(x[d]) = wx[d]$ completely escape the regularization, and as a result are essentially always included in the model.
Linear functions are \emph{not} free with saturation constraints (in fact, outside of the function $0$, they are not feasible).
When we solve~\eqref{gam} we simultaneously fit nonlinear coordinate functions while doing variable selection.

\section{Prior and related work}\label{sec:prior-related-work}

Smoothing splines also have an interpretation as the solution of an
infinite-dimensional optimization problem~\cite[\S5.4]{ESL}.
In fact, (degree-one) smoothing splines solve
\begin{equation}
\label{smoothing}
\begin{array}{ll}
\minimize & L(f) \\
\subject_to & \hat{R} (f) \le \tau,
\end{array}
\end{equation}
where \[ \label{smoothing_penalty} \hat{R}(f) = \int f'(x)^2 \; dx. \]

The solution to~\eqref{smoothing} is {\em also\/} a degree-one natural spline that saturates outside of $[0,1]$.
However, the solutions to~\eqref{smoothing} and~\eqref{problem} are {\em very\/} different.
Roughly,~\eqref{smoothing} is analogous to ridge regression, while~\eqref{problem}
 is analogous to the lasso. That is,~\eqref{smoothing} fits functions with as many knots as datapoints,
 while~\eqref{problem} often fits splines with very few knots.

Another type of spline, that is adaptive but does not saturate, are adaptive regression splines~\cite{mammen}.
These splines also arise as solutions to a functional regression problem:
 \begin{equation}
 \label{ars}
 \begin{array}{ll}
 \minimize & L(f) \\
 \subject_to & \hat{R} (f) \le \tau,
 \end{array}
 \end{equation}
 where \[ \label{arsp} \hat{R}(f) = TV(f'(x)). \]
 Note that this is~\eqref{problem} without the saturation constraint.
 Algorithms for solving~\eqref{ars} (for degree-one splines) are based on an extension of Theorem~\ref{theorem_1},
 that shows there is a solution
 to~\eqref{ars} which is actually supported on the data points
 $x_i$. Hence a  lasso algorithm can be used to find the solution.
 This also suggests a very simple method to solve our problem~\eqref{problem-nu}: we fix
 the $n$ knot points to be the values of the data $x_i$, and solve the
 finite-dimensional convex optimization problem to find the weights. While simple coordinate-descent
 methods like GLMNet~\cite{glmnet} will not immediately work because of the saturation constraint,
 they could be modified to handle the constraint. %%% We outline how this
 %%%may be done in Appendix~\ref{app:Trevor}.

 This method does work, but can be much slower than ours
 since in practice the number of knots is typically much smaller
 than $n$ for useful values of the regularization parameter $\tau$, and the finite-dimensional
 problem with $n$ basis functions is very poorly conditioned.
 With that said, the algorithm we propose --- for the piecewise linear case --- can be interpreted as a forward active set method
 for the finite dimensional problem, where we avoid explicitly evaluating all basis functions.
 One advantage of our measure-theoretic approach is that
 it immediately generalizes to higher-degree splines,
 where the support of $\mu$ need not be on data points,
 as we will see in~\S\ref{s_extensions}. In this case~\eqref{problem-nu} is truly infinite-dimensional,
 yet our algorithm can still be directly applied.

 Trend filtering is a nonparametric function estimation technique, first introduced in~\cite{trend_boyd}, that is very similar to adaptive splines.
 Indeed, as discussed in~\cite{trend_tibs}, the trend filtering estimate in the constant or piecewise-linear case is exactly the same as the adaptive spline estimate.
 Trend filtering is increasingly popular as it admits extremely efficient, robust algorithms~\cite{trend_tibs, aaditya}.
 Indeed, some of these algorithms (especially those adapted to fit GAMs~\cite{trend_gam}) may be adapted to efficiently fit saturating trend filter estimates, which would benefit from the feature selection properties of saturating splines and the computational efficiency of trend filtering.

There are a number of methods for fitting generalized additive models with spline component functions.
One approach (taken in~\cite{cosso}) is to use
the group-lasso version of~\eqref{smoothing_penalty}:
\[\label{group_lasso} R(f) = \sum_d \sqrt{\int f_d'(x)^2 \; dx}. \]
Extending this idea, \cite{chouldechova15:_gener_addit_model_selec}
use an overlap group-lasso that facilitates selection between zero, linear
and nonlinear terms.
The differences between these approaches and ours  are analogous to the differences between the standard group-lasso
and the lasso. While both do feature selection, the penalty functional~\eqref{group_lasso} does not do knot-selection within each coordinate function.

One very similar approach to fitting splines that does not require knot selection (but does not incorporate saturation) is discussed in~\cite{Rosset2007}.

\section{Examples}\label{s_examples}
In all examples we affinely preprocess the data so that all training features lie in $[0,1]$, and apply the same transformation to the test features (which thus may have values outside of $[0,1]$).
All plots are in terms of the standardized features. For the bone density and abalone datasets we select $\tau$ to minimize error on the validation sets.
For the Spam and ALS datasets we use cross-validation to estimate $\tau.$ We hold out a random subset of size 100 from the training set and train on the remaining data.
For each random validation/train split we estimate $\tau$ to minimize hold-out error and take our final estimate of $\tau$ as the mean over 50 trials.

\clearpage
\subsection{Bone density}

We start with a simple univariate dataset from~\cite[\S 5.4]{ESL}.
The response variable for this dataset is the change in spinal bone density between two doctor visits
for female adolescents as a function of age.
There are 259 data points, of which we hold out 120 for validation, leaving 139 data points to
which we fit a saturating spine.  We start with the square loss.

\begin{figure}
\begin{center}
\input{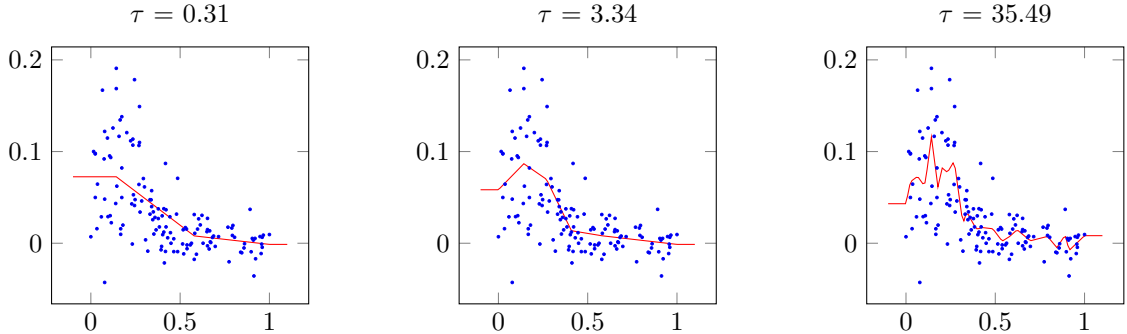}
\end{center}
\caption{Saturating splines fit to bone density data (shown as scattered points)
for 3 values of the regularization parameter $\tau$.
\emph{Top:} $\tau = 0.31$; \emph{Middle:} $\tau = 3.34$; \emph{Bottom:} $\tau = 35.45$.
 }
\label{bone-density}
\end{figure}

The results are shown in figure~\ref{bone-density}, for three
values of the regularization parameter $\tau$.
The scattered points are the training data, the solid line is the saturating
spline fit by our algorithm.
The figure demonstrates the clear link between $\tau$ and the complexity of the optimized spline.
Out-of-sample validation suggests
setting $\tau \simeq 3.34$, which achieves a validation RMSE of $0.036$.

\begin{figure}
\begin{center}
\input{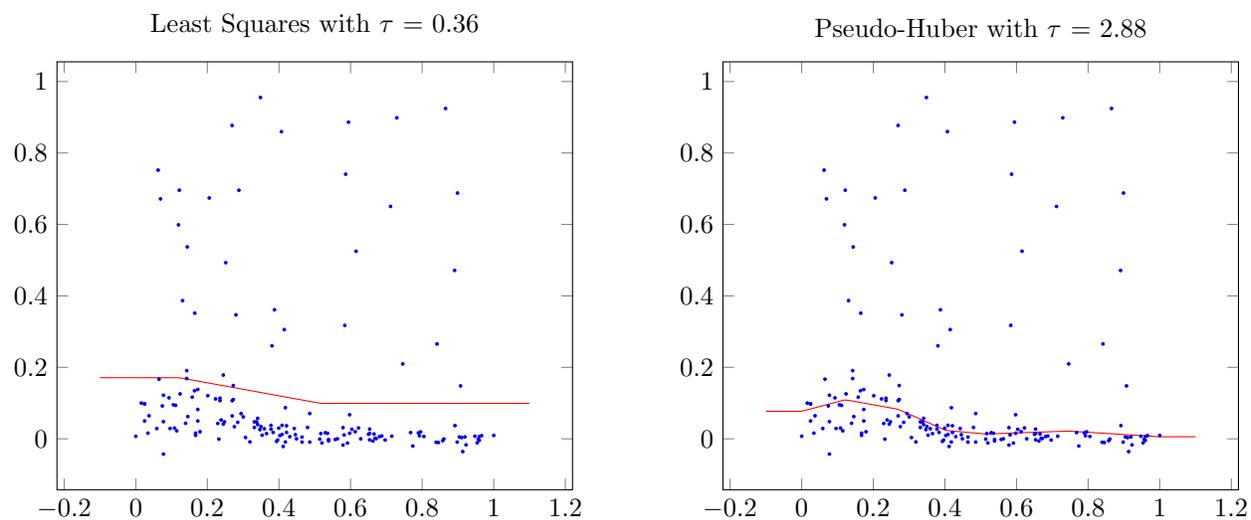}
\end{center}
\caption{Saturating splines fit to bone density data (shown as scattered points) with
simulated outliers for square loss function (left) and pseudo-Huber loss function (right),
each for the value of $\tau$ that minimizes RMSE on the test set.}
\label{bone-outliers}
\end{figure}

To demonstrate that our proposed method works with more general loss functions,
we add 30 simulated outliers to the training set and fit with the pseudo-Huber loss~\cite{pseudohuber},
a smooth approximation to the Huber loss function given by
\[l_\delta(u) = \delta \left(\sqrt{1 + \frac{u^2}{\delta}} - 1\right),\]
where $\delta > 0$ is a parameter that interpolates between the absolute value loss
and the squared loss.  For our experiment we take $\delta=0.0015$; roughly speaking, the transition
between square and linear loss occurs around $\sqrt \delta = 0.039$.
The results are shown in figure~\ref{bone-outliers}.
These plots demonstrate that our algorithm can fit losses other than the
square loss, and confirms that the pseudo-Huber loss is far more robust to outliers than
the basic square loss function.
Indeed, on the validation set the least-squares fit achieves a minimum RMSE of $0.096$,
while the pseudo-Huber fit achieves $0.038$, only slightly worse than the fit
obtained before the outliers were added to the training data.
  While this one-dimensional problem is very easy, it shows one advantage of
  the adaptive spline penalty over smoothing splines: the optimal model has only $5$ knot points.
\clearpage
\subsection{Abalone}
We fit a generalized additive model with saturating spline coordinate functions to
the Abalone dataset from the UCI Machine Learning Repository~\cite{UCI}.
The data consists of 4177 observations of 8 features of abalone along with the target variable,
the age of the abalone.
We hold out 400 data points as a validation set, leaving 3777 data points to fit the model.
The first feature (labeled sex) has three values: Male, Female, and Juvenile, which are coded with values $0,1,2$; the other 7 are (directly) real numbers.
The task is to estimate the age of the abalone from the features.

Cross-validation suggests
we choose $\tau \simeq 200$, which achieves a validation set RMSE of $2.131$.
Because the number of features is low, we can plot the entire generalized additive model.
\begin{figure}
\begin{center}
\pgfplotsset{ticks=none, xmin=-0.1, xmax=1.1, ymin=-15, ymax=15, group style={horizontal sep = 1cm}, width = 4cm}

\subfloat[$ \tau = 20$]{
\input{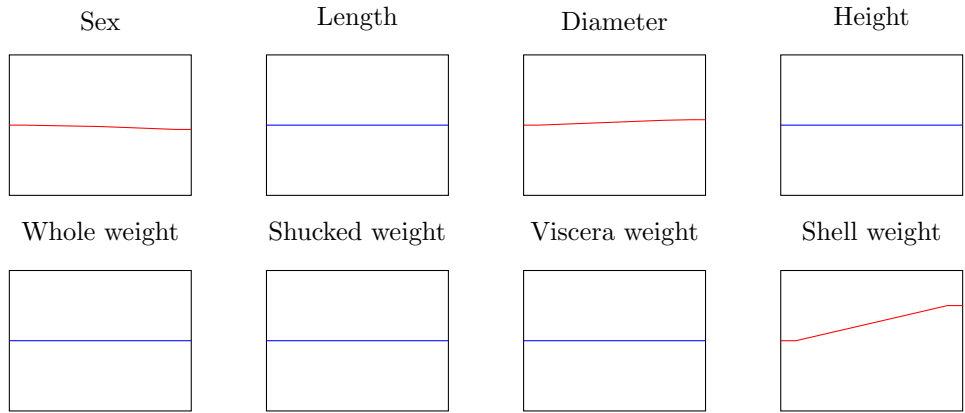}
}

\subfloat[$ \tau = 200$]{
\input{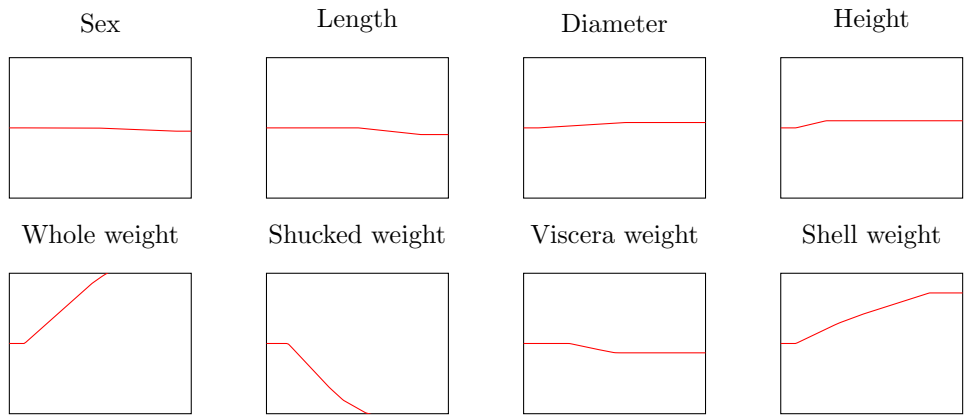}
}

\subfloat[$ \tau = 2000$]{
\input{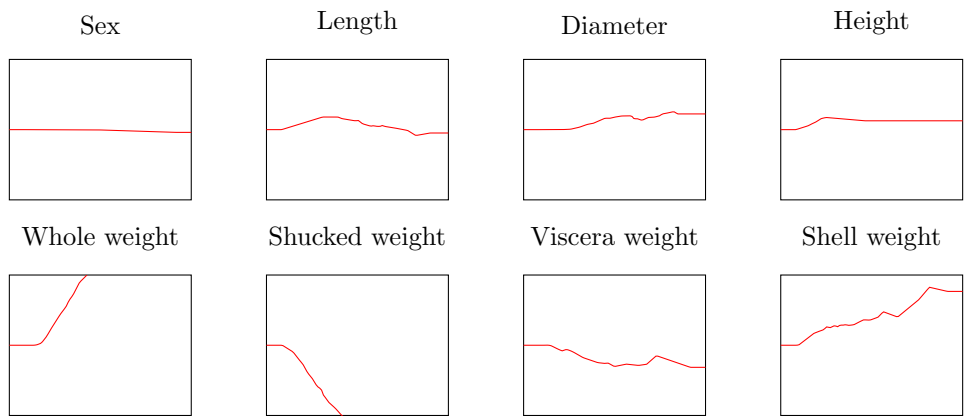}
}
\end{center}
\caption{Coordinate functions for saturating spline generalized additive models
fit to Abalone data for three values of the regularization parameter $\tau$.}
\label{abalone}
\end{figure}
Each plot shows one coordinate function $f_d$ for $d = 1, \ldots, 8$ as a function of the
standardized feature in $[0,1]$.
The coordinate functions are shown for three values of $\tau$, with the middle one
corresponding to the value that minimizes cross-validation RMSE.
When a coordinate function is zero, which means that the feature
is not used in the model, it is shown in blue.
We can see that in the case of strong regularization ($\tau =20$), several coordinates
are not used; for the best model ($\tau=200$), all features
are used, with a few having only a small effect.
It is interesting to see how the sex factors into the optimal model.
It is neutral on Male or Female, but subtracts a small
fixed amount from its age prediction for a Juvenile abalone.

This dataset is small enough that we can compare against standard adaptive splines fit using a coarse grid of $[0,1]$.
For this experiment, we fit a GAM with standard adaptive spline component functions using GLMNET~\cite{glmnet}.
The standard adaptive GAM fit, which does no variable selection, achieves a validation set RMSE of $2.137$, not significantly worse than the saturating spline model.
Our algorithm, however, selects many fewer knot points. The increased number of knots when fitting with GLMNET is perhaps due to the poor conditioning of the gridded problem.
\clearpage

\subsection{Spam}
We consider the problem of
classifying email into spam/not spam, with a dataset taken from
ESL~\cite{ESL}.
The dataset consists of 57 word-frequency features from 4601
email messages, along with their labels as spam or not spam.
Following the approach in ESL~\cite{ESL} we log-transform the features
and use the standard train/validation split,
with a training set of size 3065, and test set with 1536 samples.
We fit a saturating spline generalized additive model with standard
logistic loss.

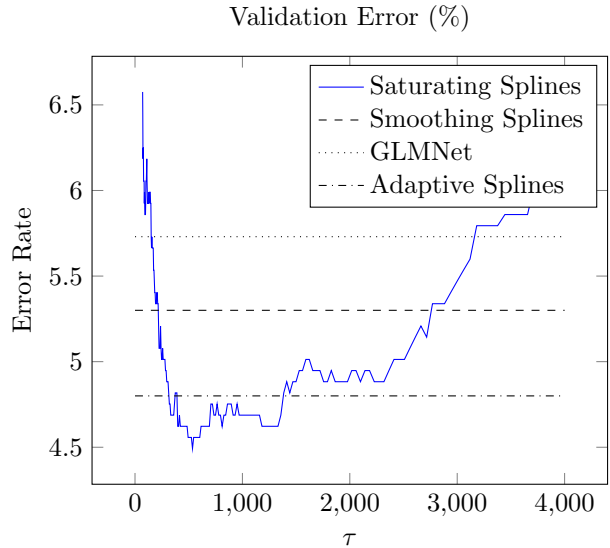
\begin{figure}
	\begin{center}
\begin{tikzpicture}[]
\begin{axis}[title = {Validation Error (\%)}, xlabel = {$\tau$}, ylabel = {Error Rate}, view = {{0}{90}}]
  \addplot+ [mark = {none}]coordinates {
(71.02594233580731, 6.575520833333333)
(72.44646118252346, 6.4453125)
(73.89539040617393, 6.25)
(75.3732982142974, 6.184895833333334)
(76.88076417858336, 6.25)
(78.41837946215503, 6.184895833333334)
(79.98674705139813, 6.0546875)
(81.58648199242609, 6.0546875)
(83.21821163227462, 6.0546875)
(84.88257586492011, 6.0546875)
(86.58022738221852, 5.924479166666666)
(88.31183192986289, 5.989583333333334)
(90.07806856846015, 5.924479166666666)
(91.87962993982936, 5.859375)
(93.71722253862595, 5.859375)
(95.59156698939847, 5.859375)
(97.50339832918644, 5.859375)
(99.45346629577017, 5.924479166666666)
(101.44253562168558, 6.0546875)
(103.47138633411929, 6.0546875)
(105.54081406080168, 6.0546875)
(107.65163034201771, 6.119791666666666)
(109.80466294885807, 6.184895833333334)
(112.00075620783522, 6.0546875)
(114.24077133199194, 5.989583333333334)
(116.52558675863177, 5.924479166666666)
(118.85609849380441, 5.924479166666666)
(121.2332204636805, 5.924479166666666)
(123.65788487295411, 5.924479166666666)
(126.1310425704132, 5.924479166666666)
(128.65366342182148, 5.989583333333334)
(131.22673669025792, 5.924479166666666)
(133.85127142406307, 5.989583333333334)
(136.52829685254434, 5.989583333333334)
(139.25886278959524, 5.989583333333334)
(142.04404004538713, 5.989583333333334)
(144.88492084629488, 5.924479166666666)
(147.78261926322077, 5.924479166666666)
(150.73827164848518, 5.729166666666666)
(153.7530370814549, 5.6640625)
(156.828097823084, 5.729166666666666)
(159.96465977954568, 5.6640625)
(163.1639529751366, 5.6640625)
(166.42723203463933, 5.6640625)
(169.75577667533213, 5.6640625)
(173.15089220883877, 5.533854166666666)
(176.61391005301556, 5.533854166666666)
(180.14618825407587, 5.46875)
(183.7491120191574, 5.403645833333334)
(187.42409425954054, 5.403645833333334)
(191.17257614473135, 5.338541666666666)
(194.99602766762598, 5.338541666666666)
(198.8959482209785, 5.338541666666666)
(202.8738671853981, 5.403645833333334)
(206.93134452910607, 5.403645833333334)
(211.0699714196882, 5.338541666666666)
(215.29137084808195, 5.338541666666666)
(219.5971982650436, 5.208333333333334)
(223.98914223034447, 5.078125)
(228.46892507495136, 5.078125)
(233.0383035764504, 5.078125)
(237.69906964797943, 5.208333333333334)
(242.45305104093902, 5.078125)
(247.3021120617578, 5.013020833333334)
(252.24815430299296, 5.013020833333334)
(257.29311738905284, 5.078125)
(262.4389797368339, 5.013020833333334)
(267.6877593315706, 5.013020833333334)
(273.041514518202, 5.013020833333334)
(278.5023448085661, 5.013020833333334)
(284.0723917047374, 4.947916666666666)
(289.75383953883215, 4.947916666666666)
(295.5489163296088, 4.8828125)
(301.45989465620096, 4.8828125)
(307.489092549325, 4.8828125)
(313.6388744003115, 4.817708333333334)
(319.91165188831775, 4.752604166666666)
(326.3098849260841, 4.752604166666666)
(332.8360826246058, 4.6875)
(339.49280427709795, 4.6875)
(346.2826603626399, 4.6875)
(353.2083135698927, 4.6875)
(360.2724798412906, 4.6875)
(367.4779294381164, 4.752604166666666)
(374.8274880268787, 4.817708333333334)
(382.3240377874163, 4.817708333333334)
(389.9705185431646, 4.817708333333334)
(397.7699289140279, 4.622395833333334)
(405.72532749230845, 4.622395833333334)
(413.8398340421546, 4.6875)
(422.11663072299774, 4.622395833333334)
(430.5589633374577, 4.622395833333334)
(439.1701426042069, 4.622395833333334)
(447.953545456291, 4.622395833333334)
(456.91261636541685, 4.622395833333334)
(466.0508686927252, 4.622395833333334)
(475.3718860665797, 4.622395833333334)
(484.8793237879113, 4.622395833333334)
(494.5769102636695, 4.557291666666666)
(504.4684484689429, 4.557291666666666)
(514.5578174383218, 4.557291666666666)
(524.8489737870882, 4.557291666666666)
(535.34595326283, 4.4921875)
(546.0528723280866, 4.557291666666666)
(556.9739297746482, 4.557291666666666)
(568.1134083701412, 4.557291666666666)
(579.475676537544, 4.557291666666666)
(591.0651900682949, 4.557291666666666)
(602.8864938696609, 4.557291666666666)
(614.944223747054, 4.622395833333334)
(627.2431082219952, 4.622395833333334)
(639.7879703864351, 4.622395833333334)
(652.5837297941638, 4.622395833333334)
(665.635404390047, 4.622395833333334)
(678.948112477848, 4.622395833333334)
(692.527074727405, 4.622395833333334)
(706.3776162219531, 4.752604166666666)
(720.5051685463922, 4.752604166666666)
(734.91527191732, 4.6875)
(749.6135773556664, 4.6875)
(764.6058489027797, 4.752604166666666)
(779.8979658808353, 4.6875)
(795.495925198452, 4.6875)
(811.4058437024211, 4.622395833333334)
(827.6339605764695, 4.6875)
(844.1866397879988, 4.6875)
(861.0703725837589, 4.752604166666666)
(878.291780035434, 4.752604166666666)
(895.8576156361428, 4.752604166666666)
(913.7747679488657, 4.6875)
(932.050263307843, 4.6875)
(950.6912685739999, 4.752604166666666)
(969.7050939454799, 4.6875)
(989.0991958243895, 4.6875)
(1008.8811797408773, 4.6875)
(1029.0588033356948, 4.6875)
(1049.6399794024087, 4.6875)
(1070.632778990457, 4.6875)
(1092.045434570266, 4.6875)
(1113.8863432616713, 4.6875)
(1136.1640701269048, 4.6875)
(1158.8873515294429, 4.6875)
(1182.0650985600319, 4.622395833333334)
(1205.7064005312325, 4.622395833333334)
(1229.8205285418571, 4.622395833333334)
(1254.4169391126943, 4.622395833333334)
(1279.5052778949482, 4.622395833333334)
(1305.0953834528473, 4.622395833333334)
(1331.1972911219043, 4.622395833333334)
(1357.8212369443424, 4.6875)
(1384.9776616832294, 4.817708333333334)
(1412.677214916894, 4.8828125)
(1440.9307592152318, 4.817708333333334)
(1469.7493743995365, 4.8828125)
(1499.144361887527, 4.8828125)
(1529.1272491252778, 4.947916666666666)
(1559.7097941077834, 4.947916666666666)
(1590.9039899899392, 5.013020833333334)
(1622.722069789738, 5.013020833333334)
(1655.176511185533, 4.947916666666666)
(1688.2800414092437, 4.947916666666666)
(1722.0456422374286, 4.947916666666666)
(1756.4865550821773, 4.8828125)
(1791.6162861838209, 4.8828125)
(1827.4486119074973, 4.947916666666666)
(1863.9975841456471, 4.8828125)
(1901.27753582856, 4.8828125)
(1939.3030865451312, 4.8828125)
(1978.089148276034, 4.8828125)
(2017.6509312415546, 4.947916666666666)
(2058.003949866386, 4.947916666666666)
(2099.1640288637136, 4.8828125)
(2141.147309440988, 4.947916666666666)
(2183.970255629808, 4.947916666666666)
(2227.649660742404, 4.8828125)
(2272.202653957252, 4.8828125)
(2317.6467070363974, 4.8828125)
(2363.9996411771253, 4.947916666666666)
(2411.279634000668, 5.013020833333334)
(2459.5052266806815, 5.013020833333334)
(2508.695331214295, 5.013020833333334)
(2558.869237838581, 5.078125)
(2610.0466225953523, 5.143229166666666)
(2662.2475550472595, 5.208333333333334)
(2715.492506148205, 5.143229166666666)
(2769.802356271169, 5.338541666666666)
(2825.1984033965928, 5.338541666666666)
(2881.7023714645247, 5.338541666666666)
(2939.3364188938153, 5.403645833333334)
(2998.123147271692, 5.46875)
(3058.0856102171256, 5.533854166666666)
(3119.247322421468, 5.598958333333334)
(3181.6322688698974, 5.794270833333334)
(3245.2649142472956, 5.794270833333334)
(3310.1702125322417, 5.794270833333334)
(3376.3736167828865, 5.794270833333334)
(3443.9010891185444, 5.859375)
(3512.7791109009154, 5.859375)
(3583.034693118934, 5.859375)
(3654.6953869813124, 5.859375)
(3727.789294720939, 6.0546875)
};
\addlegendentry{Saturating Splines};
\addplot[mark=none, black, style=dashed, samples=2, domain=0:4000]{5.3};
\addlegendentry{Smoothing Splines};
\addplot[mark=none, black, style=dotted, samples=2, domain=0:4000]{5.73};
\addlegendentry{GLMNet};
\addplot[mark=none, black, style=dashdotted, samples=2, domain=0:4000]{4.8};
\addlegendentry{Adaptive Splines};
\end{axis}

\end{tikzpicture}
\caption{Validation error for saturating spline generalized additive model
fit to Spam dataset versus regularization parameter $\tau$.}
\label{spam-error}
\end{center}
\end{figure}

Figure~\ref{spam-error} shows the validation error versus the
regularization parameter $\tau$. Cross-validation suggests the choice $\tau \simeq 1100$.
To show the benefit of nonlinear coordinate functions,
we also include the best validation error achieved using a linear model
(fit using GLMNet~\cite{glmnet}).

\begin{figure}
	\begin{center}
\input{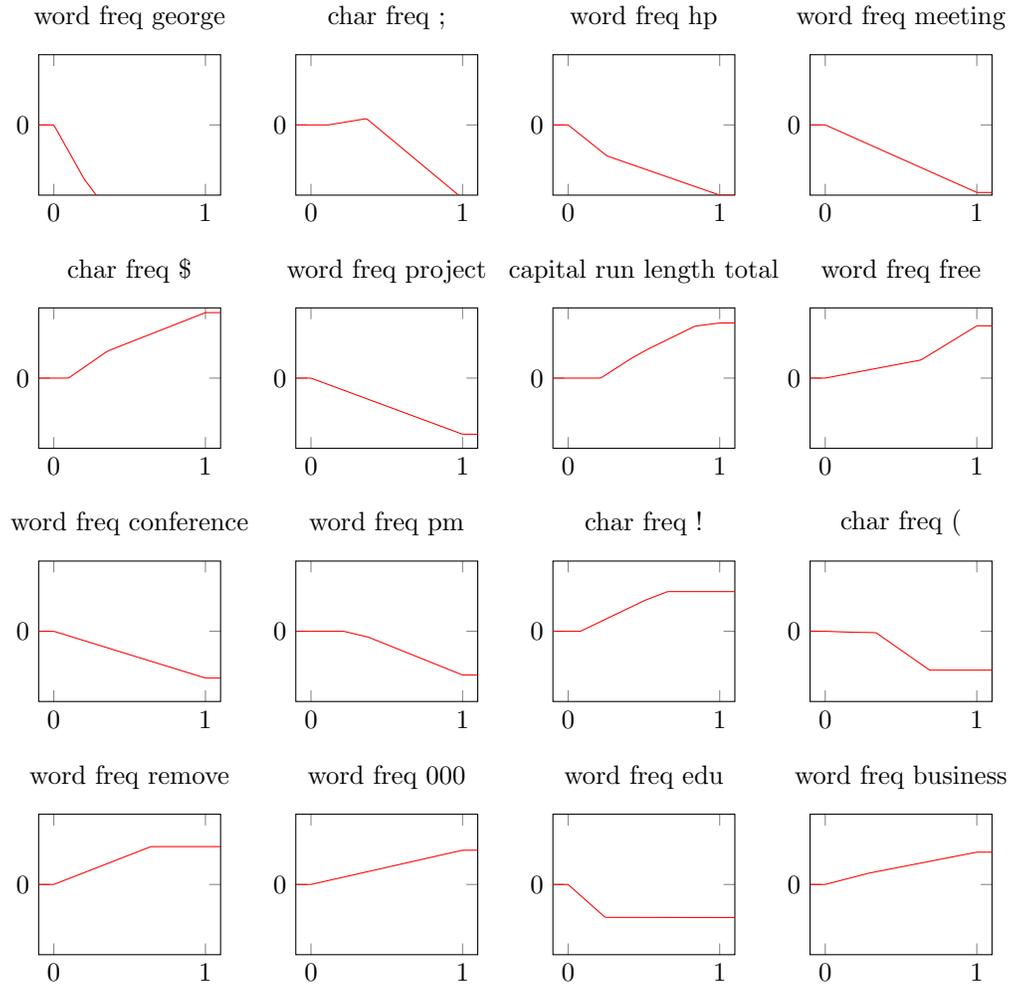}
\caption{16 coordinate functions for $\tau=500$,
labeled with the corresponding feature name. }
\label{spam-coords}
\end{center}
\end{figure}

With regularization parameter $\tau=500$, the model selects 55 of the
57 features.
We note that our saturating spline generalized additive model modestly
outperforms many methods from ESL~\cite{ESL}; for example,
smoothing splines yield 5.3\% error, while our model has an error rate
well below 5\%.
Figure~\ref{spam-coords} shows (some of) the coordinate functions for the
model with $\tau=500$. The coordinate functions use very few knots, making them readily interpretable.

For comparison, we fit a GAM with standard adaptive spline coordinate functions.
To do so, we grid each dimension with 20 knots and solve the resulting finite-dimensional problem with GLMNET~\cite{glmnet}.
Note that adaptive splines do not penalize linear functions, so there is no feature selection.
Adaptive splines achieve a minimum error of $4.8\%$, significantly worse than saturating splines.
\clearpage

\subsection{ALS}
Using this dataset we try to predict the rate of progression of ALS (amyotrophic lateral sclerosis)
in medical patients, as measured by the rate of change in their functional rating score, a measurement of functional impairment.
The dataset is split into a training set of $1197$ examples and a validation set of
$625$ additional patients. Each datapoint has dimension $369$. We fit a generalized additive
model with saturating spline component functions to the data using a least-squares objective function.
Following~\cite[\S 17.2]{computer_age}, we measure performance using mean-squared error.

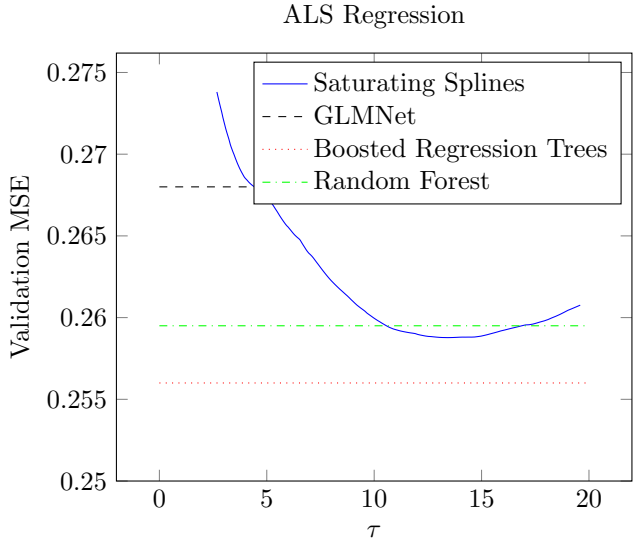
\begin{figure}
\begin{center}
\pgfplotsset{y tick label style={/pgf/number format/precision=3}}
\begin{tikzpicture}[]
\begin{axis}[title = {ALS Regression}, xlabel = {$\tau$}, ylabel = {Validation MSE}, view = {{0}{90}}, ytick={0.25, 0.255, 0.26, 0.265, 0.27, 0.275}, yticklabel style={/pgf/number format/fixed,
                  /pgf/number format/precision=3}, ymin=0.25]\addplot+ [mark = {none}]coordinates {
(2.678033494476762, 0.27380602840594787)
(2.7048138294215294, 0.27364927244014864)
(2.7318619677157447, 0.2734929627665019)
(2.759180587392902, 0.27333744876814087)
(2.786772393266831, 0.2731848069533637)
(2.8146401171994992, 0.273032982239276)
(2.8427865183714944, 0.27288398527494706)
(2.8712143835552095, 0.27273598642111585)
(2.8999265273907615, 0.27258699354367194)
(2.928925792664669, 0.2724153625530306)
(2.9582150505913156, 0.2722389397229614)
(2.9877972010972287, 0.2720849098086779)
(3.017675173108201, 0.27192779856610655)
(3.0478519248392835, 0.2717755106776719)
(3.078330444087676, 0.27161303544746446)
(3.109113748528553, 0.27146829000164496)
(3.1402048860138385, 0.2713291553240145)
(3.171606934873977, 0.2711933319969454)
(3.203323004222717, 0.27106004474214035)
(3.2353562342649442, 0.2709227894663723)
(3.2677097966075936, 0.27076580492204855)
(3.3003868945736694, 0.27060067939237686)
(3.333390763519406, 0.27046585605998735)
(3.3667246711546, 0.27034767411390326)
(3.400391917866146, 0.27022535089459276)
(3.4343958370448076, 0.2701098599653041)
(3.4687397954152557, 0.27000107488341335)
(3.5034271933694083, 0.26987845909989716)
(3.5384614653031026, 0.269751052254566)
(3.573846079956134, 0.269633519268216)
(3.6095845407556952, 0.26951359347151976)
(3.645680386163252, 0.2693978917173596)
(3.6821371900248847, 0.26929186092502055)
(3.7189585619251337, 0.2691818320187)
(3.756148147544385, 0.2690795271845665)
(3.793709629019829, 0.2689819503940225)
(3.8316467253100273, 0.2688870785190625)
(3.8699631925631275, 0.26879549349686876)
(3.9086628244887587, 0.2687037076431403)
(3.947749452733646, 0.2686137575943689)
(3.9872269472609827, 0.2685437021108123)
(4.027099216733593, 0.2684868407051424)
(4.067370208900929, 0.2684319399121338)
(4.108043910989938, 0.26837353743957115)
(4.149124350099838, 0.2683073206429823)
(4.190615593600836, 0.2682516723823155)
(4.232521749536844, 0.26819905200104305)
(4.274846967032213, 0.26814901562198423)
(4.317595436702534, 0.2681022419751077)
(4.36077139106956, 0.26805934304297646)
(4.404379104980255, 0.2680192055699368)
(4.448422896030058, 0.2679802316132091)
(4.492907124990358, 0.26794373108523983)
(4.537836196240262, 0.2678991063567618)
(4.583214558202664, 0.2678413681807823)
(4.629046703784691, 0.26778992032398735)
(4.675337170822537, 0.2677407709413634)
(4.722090542530763, 0.26768941122532547)
(4.7693114479560705, 0.2676030535524502)
(4.817004562435631, 0.26754292375277683)
(4.865174608059988, 0.26748806336176895)
(4.913826354140587, 0.2674324180767755)
(4.962964617681993, 0.26738169416739993)
(5.012594263858813, 0.2673232116163466)
(5.062720206497401, 0.2672652674095153)
(5.113347408562375, 0.2672105488131549)
(5.164480882647999, 0.2671250724726071)
(5.216125691474479, 0.2670228892995172)
(5.268286948389224, 0.26692028509795546)
(5.320969817873116, 0.26681790573541614)
(5.374179516051847, 0.26670725546424157)
(5.427921311212366, 0.266598356665142)
(5.482200524324489, 0.26649079859498265)
(5.537022529567734, 0.26637320376072887)
(5.592392754863411, 0.2662528543335984)
(5.648316682412045, 0.26613883979257624)
(5.704799849236166, 0.2660271148370283)
(5.761847847728528, 0.26591746614976625)
(5.819466326205813, 0.26580964016869757)
(5.877660989467871, 0.2657061193283091)
(5.9364375993625496, 0.2656139506943721)
(5.9958019753561755, 0.2655377262957106)
(6.055759995109737, 0.2654399838867944)
(6.116317595060834, 0.26534405798977534)
(6.177480771011442, 0.26524324151889034)
(6.239255578721557, 0.2651445768803806)
(6.301648134508772, 0.26505456126601984)
(6.36466461585386, 0.2649750284023297)
(6.428311262012398, 0.26489901241064956)
(6.492594374632523, 0.2648318325550966)
(6.557520318378848, 0.2647392323567683)
(6.623095521562637, 0.26459232310839487)
(6.689326476778263, 0.2644507869991721)
(6.7562197415460465, 0.2643137733736448)
(6.823781938961507, 0.2641719042136044)
(6.892019758351122, 0.2640444323595675)
(6.960939955934633, 0.26393604699435397)
(7.03054935549398, 0.26385370239603384)
(7.100854849048919, 0.2637586058873736)
(7.171863397539409, 0.26364617573430654)
(7.243582031514803, 0.2635186289009223)
(7.316017851829951, 0.26339020771380567)
(7.38917803034825, 0.26326482711557103)
(7.463069810651733, 0.2631347965519178)
(7.537700508758251, 0.263003808541057)
(7.613077513845833, 0.2628738083945399)
(7.689208288984291, 0.26274539414863873)
(7.766100371874135, 0.2626196795781744)
(7.843761375592877, 0.26249563960429056)
(7.922198989348805, 0.2623745999732193)
(8.001420979242294, 0.2622595305105726)
(8.081435189034718, 0.26215135881232743)
(8.162249540925066, 0.26204293919996907)
(8.243872036334317, 0.2619377463206055)
(8.32631075669766, 0.26183348408019774)
(8.409573864264637, 0.2617292265378524)
(8.493669602907284, 0.26163262845229635)
(8.578606298936357, 0.26153120357207404)
(8.66439236192572, 0.261428750091164)
(8.751036285544979, 0.2613279109553749)
(8.83854664840043, 0.26121720956981387)
(8.926932114884433, 0.26109896544949784)
(9.016201436033278, 0.26098897902013724)
(9.10636345039361, 0.26087487927410835)
(9.197427084897546, 0.26076479215358156)
(9.289401355746522, 0.2606695932384654)
(9.382295369303987, 0.26057579991357216)
(9.476118322997028, 0.26048696611892797)
(9.570879506226998, 0.26039197260835356)
(9.666588301289268, 0.26027366019219933)
(9.763254184302161, 0.2601818149548904)
(9.860886726145182, 0.2600912455746421)
(9.959495593406634, 0.2599967610084157)
(10.0590905493407, 0.2599085490161833)
(10.159681454834107, 0.25982294273547507)
(10.261278269382448, 0.25974315954462285)
(10.363891052076273, 0.2596589148178964)
(10.467529962597036, 0.25957806889743046)
(10.572205262223006, 0.2594984333784994)
(10.677927314845236, 0.2594230946922549)
(10.78470658799369, 0.2593619909124888)
(10.892553653873627, 0.25930419543690764)
(11.001479190412363, 0.25926353698766036)
(11.111493982316487, 0.25922205478074084)
(11.222608922139653, 0.2591821494970956)
(11.334835011361049, 0.25915075237325486)
(11.44818336147466, 0.2591213291585943)
(11.562665195089407, 0.2590933027554475)
(11.6782918470403, 0.2590675910734846)
(11.795074765510703, 0.25904636355010285)
(11.91302551316581, 0.2590210508566062)
(12.032155768297468, 0.25896832556127686)
(12.152477325980442, 0.25893219573190696)
(12.274002099240246, 0.25890118931866307)
(12.396742120232648, 0.2588713014934269)
(12.520709541434975, 0.2588484033368842)
(12.645916636849325, 0.25883938629058556)
(12.772375803217818, 0.258821783484551)
(12.900099561249997, 0.2588051076661256)
(13.029100556862497, 0.2587909876173615)
(13.159391562431122, 0.25878225972817515)
(13.290985478055433, 0.2587790700648265)
(13.423895332835986, 0.25877550836376156)
(13.558134286164346, 0.2587784962711553)
(13.69371562902599, 0.2587830159922836)
(13.83065278531625, 0.25879109606743594)
(13.968959313169412, 0.2588006072550138)
(14.108648906301106, 0.2588055022452189)
(14.249735395364118, 0.25880422032248895)
(14.39223274931776, 0.25880696113234636)
(14.536155076810937, 0.258805746049019)
(14.681516627579047, 0.25881346993981263)
(14.828331793854836, 0.25883891801105974)
(14.976615111793384, 0.2588664961384613)
(15.126381262911318, 0.25891107843315947)
(15.27764507554043, 0.2589588132986296)
(15.430421526295834, 0.2590140880937166)
(15.584725741558794, 0.25906770795677114)
(15.740572998974383, 0.25912398825202493)
(15.897978728964127, 0.25918572494622927)
(16.05695851625377, 0.259234867098459)
(16.217528101416306, 0.25928886086596387)
(16.37970338243047, 0.25934269944622823)
(16.543500416254776, 0.25939260599960784)
(16.708935420417323, 0.25944602135620864)
(16.876024774621495, 0.2595010989906039)
(17.04478502236771, 0.25955176074362235)
(17.215232872591386, 0.2595640599501299)
(17.3873852013173, 0.25960212862611487)
(17.561259053330474, 0.259666083988255)
(17.73687164386378, 0.25973631467300873)
(17.91424036030242, 0.25979921394972644)
(18.093382763905442, 0.2598804565332245)
(18.274316591544498, 0.25998508621906996)
(18.45705975745994, 0.2600818827331386)
(18.64163035503454, 0.2601907934985869)
(18.828046658584885, 0.2603136395324247)
(19.016327125170733, 0.26044360545859824)
(19.206490396422442, 0.26054586665046053)
(19.398555300386665, 0.26065293585061006)
(19.59254085339053, 0.26076131234312566)
};
\addlegendentry{Saturating Splines};
\addplot[mark=none, black, style=dashed, samples=2, domain=0:20]{0.268};
\addlegendentry{GLMNet};
\addplot[mark=none, red, style=dotted, samples=2, domain=0:20]{0.256};
\addlegendentry{Boosted Regression Trees};
\addplot[mark=none, green, style=dashdotted, samples=2, domain=0:20]{0.2595};
\addlegendentry{Random Forest};
\end{axis}

\end{tikzpicture}
\end{center}
\caption{Validation MSE on ALS dataset versus regularization parameter
$\tau$.}
\label{als}
\end{figure}

We estimate the optimal value of $\tau$ using cross validation with a hold-out size of 100 examples and 50 samples; this procedure suggests $\tau = 13$.
Figure~\ref{als} shows the validation error versus the regularization
parameter $\tau$; the value of $\tau$ selected by cross validation achives low error.
On the same plot, we also show the results from~\cite{computer_age} using boosted regression trees and random forests.
The optimal saturating spline GAM model selects only 50 out of the 369 features, in contrast to boosted regression trees, which use
267. The saturating spline GAM model performs comparably to boosted regression trees and random forests. This is surprising
as the saturating spline GAM has no interaction terms. It also uses substantially fewer features, further improving interpretability.

Again we fit a GAM with standard adaptive spline coordinate functions (using GLMNET) to show the advantage of saturation.
The standard adaptive spline fit achieves an MSE of $0.547,$ substantially worse than any other model.
We speculate that this is because the unpenalized linear functions lead to immediate overfitting.
Indeed, removing the unpenalized linear functions and fitting a model with only hinges gives very similar performance to the saturating spline fit,
suggesting that the main advantage of saturation for this application is the removal of the unpenalized linear functions.

\paragraph{Practical advantages of saturating splines}
These experiments show that saturating splines achieve competitive performance on small classification and regression datasets.
In addition, the experiments demonstrate that saturating splines exhibit both knot selection and feature selection --- in the context of fitting GAMs.
While it is no surprise that saturating splines select fewer knots than smoothing splines (which choose a fully-dense set of knots), it is somewhat surprising that our algorithm
selects fewer knots than even \emph{adaptive} splines fit with GLMNET.
Finally, the Spam and ALS datasets demonstrate a major advantage of saturating splines over adaptive splines: they simultaneously perform non-linear coordinate function fitting and feature selection.
This aids in generalization performance and interpretability. In particular, for the ALS dataset saturating spline GAMs achieve \emph{half} the test MSE of adaptive spline GAMs by selecting only 50 of 369 available features.

\section{Higher-degree splines}
\label{s_higher_degree}

In the majority of this paper we focused on the functional regression
problem~\eqref{problem}, with a total variation constraint on the first derivative and
a saturation constraint on the zeroth derivative (the function itself).
In this section,
we consider constraints on higher order derivatives, which lead to solutions that are splines of higher degrees.

\begin{equation}
\label{family}
\begin{array}{ll}
\minimize & L(f) \\
\subject_to & \TV (f^{(k)}) \le \tau, \\
& f^{(k-j)}(x) = 0,~\forall x \not\in [0,1].
\end{array}
\end{equation}

We consider the family of nonparametric function estimation problems indexed by $0 \le j \le k$.
This is the analogue of the functional regression problem~\eqref{problem} with a total variation constraint on the
$k$-th derivative and a saturation constraint on the $(k-j)$-th derivative.
The saturating spline case from the rest the paper is the special case of~\eqref{family} with $k=1$, $j=0$.
Widely used cubic natural splines correspond to $k=3$, $j=1$.
Note that unlike natural splines, which are only defined for some values of $j$ and $k$, there are no constraints on $j$ and $k$.

We now show that higher-degree saturating splines solve~\eqref{family} in general.
As $f^{(k)}$ is of bounded TV, there exists a measure $\mu$ s.t.
$f^{(k)}(x) = \int 1(t \le x)\; d\mu(t)$.
Then we have

\begin{align*}
	f^{(k-j)}(x) &= \int \ldots \int f^{(k)}(x) dx \ldots dx \\
	&= \int \ldots \int \int 1(t \le x)\;d\mu(t) dx \ldots dx \\
	&= j! \int {(x - t)}^j_+ \; d\mu(t) + \sum_{l = 0}^{j-1} w_l x^l
\end{align*}
for some $w_l$. In the above, all iterated integrals take place $j$ times.

Note that the constraint that $f^{(k-j)}(x) = 0$ for all $x < 0$ implies
that the polynomial term, $\sum_{l = 0}^{j-1} w_l x^l$ is identically zero.
So, we have
\[ f^{(k-j)}(x) = j! \int {(x - t)}^j_+ \; d\mu(t).\]
For $x>1$, we can remove the nonlinearity, that is, for $x>1$,  $f^{(k-j)}(x)$
is simply the integral of a polynomial in $x$. We can pull terms involving $x$
 out of the integral to get a polynomial in $x$ whose coefficients are
 nonzero multiples of the first $j$ moments of $\mu$:
 \[ f^{(k-j)}(x) = j! \sum_{l=0}^j \binom{j}{l}x^{j-k}\int {(-t)}^k \;d\mu(t). \]

Again, we note that as this polynomial is identically zero for infinitely many
points, all of the coefficients must be zero. In terms of the measure $\mu$, this means:

\begin{align*}
\int t^l \;d\mu(t) = 0\quad \textrm{for } l = 0,\ldots, j.
\end{align*}

This shows that the constraint that the $(k-j)$-th derivative of $f$ saturate translates to
constraints on all moments of $\mu$ up to the $j$-th moment.

While the conditional gradient step becomes more complex with the addition of
more moment constraints, the approach taken in this paper can still be applied to~\eqref{family}
as long as $j$ is fairly small --- the conditional gradient step for~\eqref{family}
involves a nonconvex optimization problem over $[0,1]^{j+2}$.
This is because we need at least $(j+2)$ point-masses to satisfy the moment constraints.
So, fitting quadratic splines that saturate to linear
is very easy --- in fact the code to do so is essentially identical to that
 for fitting piecewise linear saturating splines splines --- but fitting quadratic
  splines that saturate to constant is slightly more difficult due to the additional
	linear constraint on the measure $\mu$. Unfortunately for larger values of $j$ and $k$, we can no longer hope to find the conditional gradient
  analytically and must resort to recursive gridding or other global optimization algorithms to find the locations of the new knots.

	\begin{figure}
	\centering
	\begin{tikzpicture}
	\begin{groupplot}[group style={group size= 2 by 2, horizontal sep = 2cm, vertical sep = 2cm}]
	\nextgroupplot[
	    axis lines = left,
	    xlabel = $x$,
			% ylabel = $k=2 and j=0$,
			xmax = 1.2,
			xmin = -0.2,
			ymax = 0.2,
			ymin = -0.2,
	]
	\addplot [
	    domain=-0.1:1.1,
	    samples=100,
	    color=blue,
	]
	{+max(x-0.5,0.0)^2 -max(x-0.6,0.0)^2};
	\addplot [dashed]
	   coordinates {(0.5, -0.5)  (0.5, 0.5)};
	 \addplot [dashed]
		coordinates {(0.6, -0.5)  (0.6, 0.5)};

		\nextgroupplot[
		    axis lines = left,
		    xlabel = $x$,
				% ylabel = $k=2 and j1$,
				xmax = 1.2,
				xmin = -0.2,
				ymax = 0.2,
				ymin = -0.2,
		]
		\addplot [
		    domain=-0.1:1.1,
		    samples=100,
		    color=blue,
		]
		{-max(x-0.4,0.0)^2 +0.5*max(x-0.2,0.0)^2 +0.5*max(x-0.6,0.0)^2};
		\addplot [dashed]
		   coordinates {(0.4, -0.5)  (0.4, 0.5)};
	 \addplot [dashed]
		   coordinates {(0.2, -0.5)  (0.2, 0.5)};
		 \addplot [dashed]
			coordinates {(0.6, -0.5)  (0.6, 0.5)};

			\nextgroupplot[
			    axis lines = left,
			    xlabel = $x$,
					ylabel = $\mu$,
					xmax = 1.2,
					xmin = -0.2,
					ymax = 1.5,
					ymin = -1.5,
					axis x line=center,
			]
			\addplot +[mark=triangle*,
			   mark options={scale=1},
			   scatter,ycomb,
			   visualization depends on={(y-abs(y))/abs(y-abs(y)) \as \sign},
			   scatter/@pre marker code/.code={\scope[rotate=180*\sign,yshift=-2pt]}
			   ] coordinates {(0.5, 1.0)  (0.6, -1)};

				 \nextgroupplot[
		 				axis lines = left,
		 				xlabel = $x$,
		 				ylabel = $\mu$,
		 				xmax = 1.2,
		 				xmin = -0.2,
		 				ymax = 1.5,
		 				ymin = -1.5,
		 				axis x line=center,
		 		]
				% -max(x-0.4,0.0)^2 +0.5*max(x-0.2,0.0)^2 +0.5*max(x-0.6,0.0)^2
		 		\addplot +[mark=triangle*,
		 			 mark options={scale=1},
		 			 scatter,ycomb,
		 			 visualization depends on={(y-abs(y))/abs(y-abs(y)) \as \sign},
		 			 scatter/@pre marker code/.code={\scope[rotate=180*\sign,yshift=-2pt]}
		 			 ] coordinates {(0.4, -1.0)  (0.2, 0.5) (0.6, 0.5)};
	\end{groupplot}
	\end{tikzpicture}
	\caption{The top two plots show conditional gradients for $k=2$ with $j=0$ and $j=1$ respectively.
	The dashed lines denote the locations of the point masses: when $j=1$, the conditional gradient consists of three point masses.
	The bottom plots show the corresponding measures.}\label{higher_order}
	\end{figure}
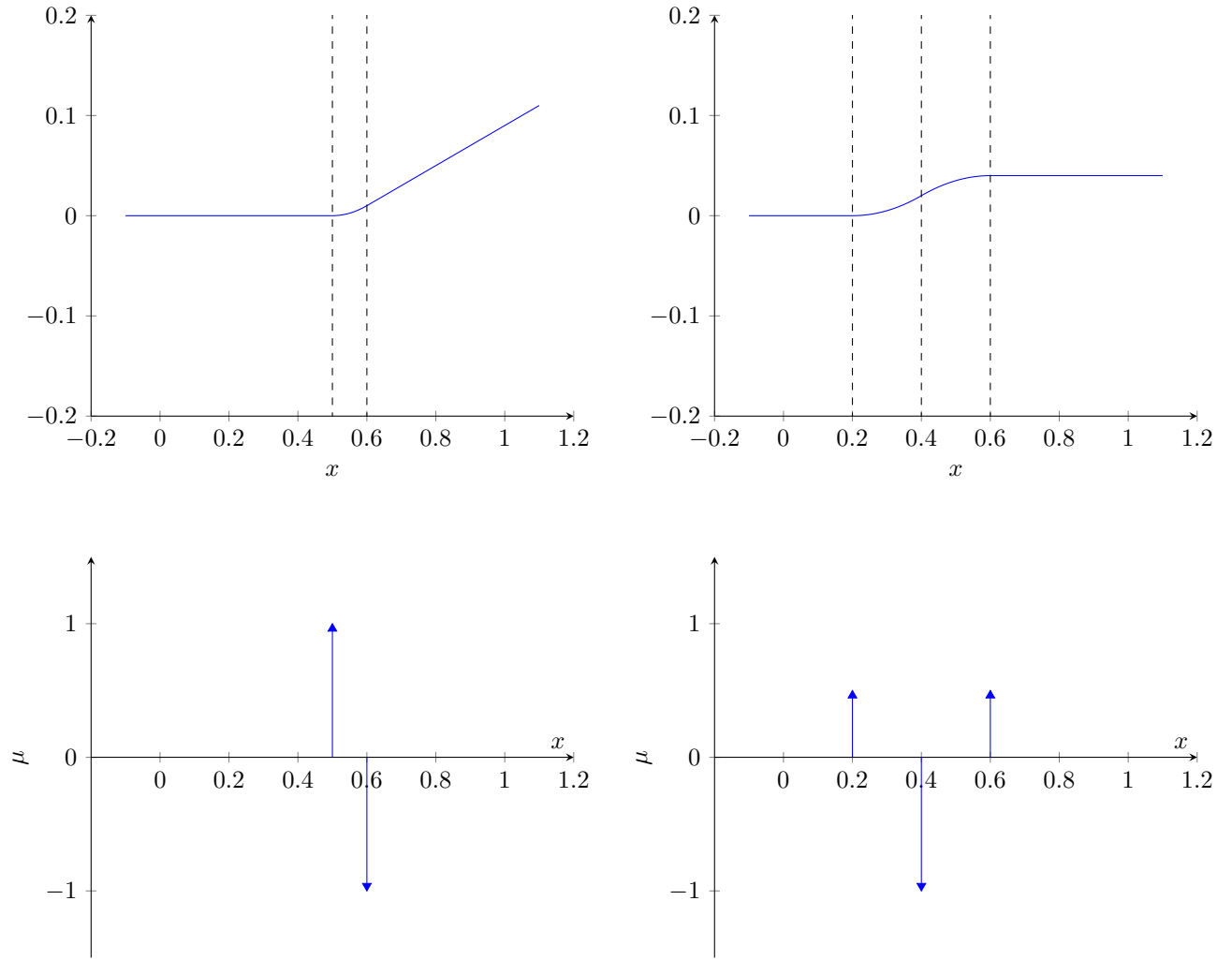
	\clearpage

\section{Variations and extensions}
\label{s_extensions}

While saturation is often a natural prior, the approach we take in this paper
can also be applied to other (convex) variations on~\eqref{problem-nu}.
For example, we could add the constraint that the fitted function is monotone
nondecreasing, or takes values in a given interval.

A simple algorithmic extension would be to incorporate nonconvex optimization in the spirit of~\cite{adcg}.
At each iteration we adjust the weights of the atomic measure ($w$), but we could also adjust the knot locations ($t$).
The objective in~\eqref{finite_sup} is nonconvex in $t_i$, but we can still attempt to find a local minimum.
As long as we do not increase the objective function the algorithm is still guaranteed to converge~\cite{adcg}.
In the case of degree one splines, we can use the fact that the knot points can,
without loss of generality, be chosen to be on the data points to make discrete adjustments to the knot locations.

To fit vector-valued functions, for example in multiclass classification, we would need to extend~\eqref{problem-nu} to use {\em vector-valued} measures.
This is the natural measure-theoretic analogue to the group-lasso.

In multivariate fitting problems with significant interactions between features
generalized additive models may underfit.
One possible solution is to use single-layer neural networks: i.e.
learn functions of the form
\[ \label{NN} x \mapsto \sum_{i=1}^K w_i(v_i^Tx - t_i)_+. \]
In the above, $v_i$ are constrained to lie in the unit ball.
Unfortunately, the conditional gradient step for networks of this form is NP-hard~\cite{francis}.
In many practical applications, however, we might expect that the degree of the interaction is bounded.
That is, each $v_i$ has bounded cardinality.
If we assume $\|v_i\|_0 \le 2$, i.e. we only fit pairwise interactions, we can still
apply the conditional gradient method. In this case, the fitting function is a sum of functions of pairs of the variables,
formed from the basis elements
\[
((\cos \theta) x_p+(\sin \theta) x_q - t)_+,
\]
with (continuous) parameters $\theta$ and $t$ and (index) parameters
$p$ and $q$ (i.e. $v = (\cos \theta)e_p + (\sin \theta)e_q$).
(This is practical only if $d$ is small enough.)
Such functions capture nonlinear relationships between (pairs of) variables.

\section{Conclusion}
In this paper we propose a modification of the adaptive spline regression model --- namely saturation constraints.
We show that saturating splines inherit knot-selection from adaptive splines, and have a very important quality
in the context of generalized additive models: feature selection. This allows saturating spline generalized additive models to remain interpretable and (crucially) avoid overfitting when applied to multivariate data.
We also propose a simple, effective algorithm based on the standard conditional gradient method for solving the saturating spline estimation problem with arbitrary convex losses.
Finally, we apply our algorithm to several datasets, demonstrating the simplicity of the resulting models.

\section*{Acknowledgements}
We would like to thank Aaditya Ramdas for many helpful discussions about trend filtering.
NB was generously supported by a Google Fellowship from the Hertz Foundation.
\printbibliography
% \clearpage
\appendix
\section{Implementation details}
We provide a simple, unoptimized implementation in the Rust language.
The runtime of our algorithm is dominated by the fully-corrective step, that is, solving the finite-dimensional convex optimization problem~\eqref{finite_sup}.
We solve~\eqref{finite_sup} using a proximal Newton method and the standard conditional gradient method with exact linesearch.
To be precise, at each iteration, we form the second-order approximation to the objective function
\[ f(w) \simeq C +  (w - \hat{w})^T\nabla_w f(\hat{w}) + \frac{1}{2} (w - \hat{w})^T \nabla^2 f(\hat{w}) (w- \hat{w}) \]
which we then minimize (over the constraint set) using the standard conditional gradient method with (exact) linesearch.
Note that this is a Newton step with fixed step-length of $1$: as in GLMNET \cite{glmnet}, we omit a line search in the interest of speed.

We chose to use a proximal Newton method because of its relative simplicity;
other standard convex optimization algorithms may give much better practical performance, especially when the number of data points, $n$, is extremely large.
\section{Saturating hinges}\label{app:Trevor}
In this section we introduce a heuristic for solving an approximation to~\eqref{problem-nu} using existing algorithms for the lasso.
\def\weight{w}
Here we consider the case where $\hat{R}(f) = TV(f'(x))$, and hence as
pointed out in Section~\ref{sec:prior-related-work} the solution is an
expansion in piecewise linear splines with knots at the unique data points.
Let $h_j$ be a hinge function at knot $t_j$: $h_j(x)=(x-t_j)_+$, and
suppose we have knots $t_1<t_t<\cdots<t_k$.
Define $f(x)=\weight_0+\sum_{j=1}^k\weight_jh_j(x)$.
Given a sample $T=\{(x_i,y_i)\}_1^N$,  solving (\ref{problem-nu})
amounts to solving
\begin{equation}
  \label{eq:1}
\minimize_{\weight_0,\weight}\ENCMIN{
\sum_{i=1}^N\ell(y_i,f(x_i))+\lambda\|\weight\|_1
}
\quad
\mbox{s.t. }f'(t_k)=0.
\end{equation}
Here we've exchanged a constraint on the total variation of $TV(f'(x))$ with a penalty.
The condition $f'(t_k)=0$ is equivalent to $\sum_{j=1}^k\weight_j=0$.
If the points $x_i$ are unique, then $k=n$; irrespective
$t_1=\min(x_i)$, $t_k=\max(x_i)$ and by construction the estimate
$\hat f$ is constant beyond the data.

Without the gradient condition $f'(t_k) = 0$, solving \eqref{eq:1} amounts to a large
lasso problem, for which efficient software is available. Here our
goal is to transform the problem to get rid of this constraint.
For more generality we  do this for an arbitrary set of ordered knots.

\begin{figure}[h]
  \centering
  \includegraphics[width=.6\textwidth]{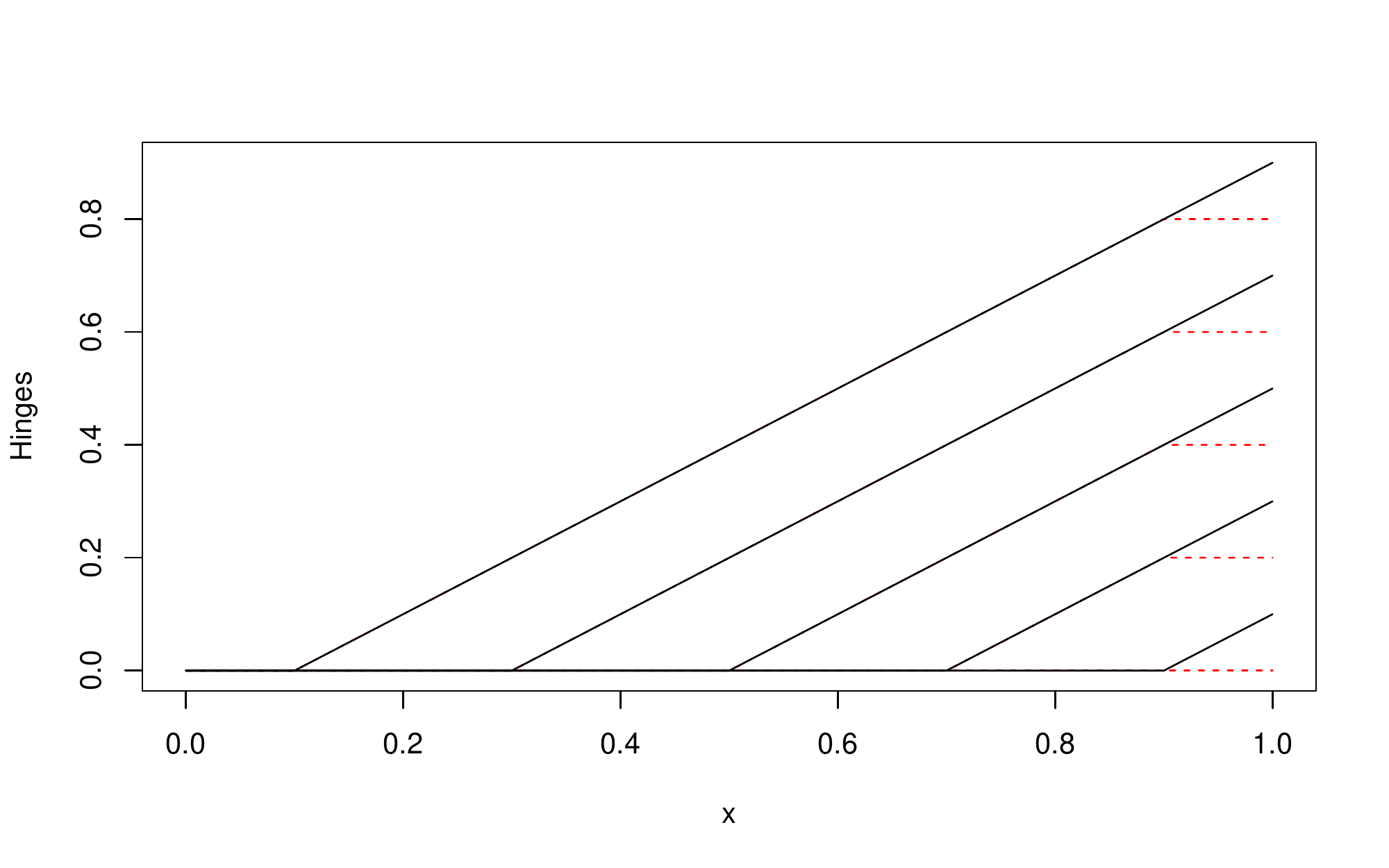}
  \caption{Hinges and saturating hinges.}
  \label{fig:one}
\end{figure}

Suppose $k<n$ and the right-most knot is inside the range of the data.
Consider the following formulation.
Let $s_j(x)=h_j(x)-h_k(x)$, a ``saturating'' hinge function. It looks
like a piecewise linear sigmoid, and goes horizontal at $t_k$ (see
figure~\ref{fig:one}).

Without the $\ell_1$ constraint (or when $\lambda=0$) the solution to
\eqref{eq:1} is equivalent to the solution to the problem with the
reduced basis
$g(x)=\theta_0+\sum_{j=1}^{k-1}\theta_js_j(x)$:
\begin{equation}
  \label{eq:2}
  \minimize_{\theta_0,\theta}\sum_{i=1}^N\ell(y_i,g(x_i)).
\end{equation}
This is easy to see. $f$ is an affine expansion in the $h_j$, and
hence any nonsingular $k\times k$ transformation $C$  of the vector of
functions $h(x)=(h_1(x),h_2(x),\ldots,h_k(x))$ spans the same space.
It is easy to see that with $s(x)=C^Th(x)$, and
$$C=\left[
  \begin{array}{rrrrr}
    1&0&\cdots&0&0\\
    0&1&\cdots&0&0\\
    \vdots&\vdots&\ddots&\vdots&\vdots\\
    0&0&\cdots&1&0\\
    -1&-1&\cdots&-1&-1
  \end{array}
\right]
$$
that $s_j(x)$ are as described (and $s_k(x)=-h_k(x)$).
Now $h(x)^T\weight=h(x)^TCC^{-1}\weight=s(x)^T\theta$, with
$\theta=C^{-1}\weight$. However, in this case $C^{-1}=C$,
and hence $\theta_j=\weight_j,\;j=1,\ldots,k-1$ and
$\theta_k=\sum_{j=1}^k\weight_j$. In this new basis, imposing the
constraint amounts to setting $\theta_k=0$, or simply deleting the
last basis function. So fitting the linear model $f$
subject to $f'(t_k)=0$ is equivalent to fitting the model $g$ without
constraints, and in fact the $\theta_j=\weight_j,\;j=1,\ldots,k-1$.

So in summary,  fitting a constrained optimization with the hinge functions is
equivalent to fitting an unconstrained optimization with the reduced
set of saturating hinges. The remaining question
is does this also work
with the penalty $\lambda\|\weight\|$ as in \eqref{eq:1}.
Not quite, but close. It turns out we are still missing a penalty term
$\lambda|\sum_{j=1}^{k-1}\theta_j|$.
Hence the transformed problem is
\begin{equation}
  %\label{eq:1}
\minimize_{\theta_0,\theta\in\reals^{k-1}}\ENCMIN{
\sum_{i=1}^N\ell(y_i,g(x_i))+\lambda\|\theta\| +\lambda|\mbox{$\sum_{j=1}^{k-1}\theta_j$}|.
}
\end{equation}
If we are willing to ignore this last penalty, we can fit the
saturated spline model using a fast lasso solver, such as {\tt
  glmnet}. By generating such a basis for each variable in a GAM, this
same approach can be used to fit a saturated GAM regularization path.

\begin{figure}[hbtp]
  \centering
  \includegraphics[width=.5\textwidth]{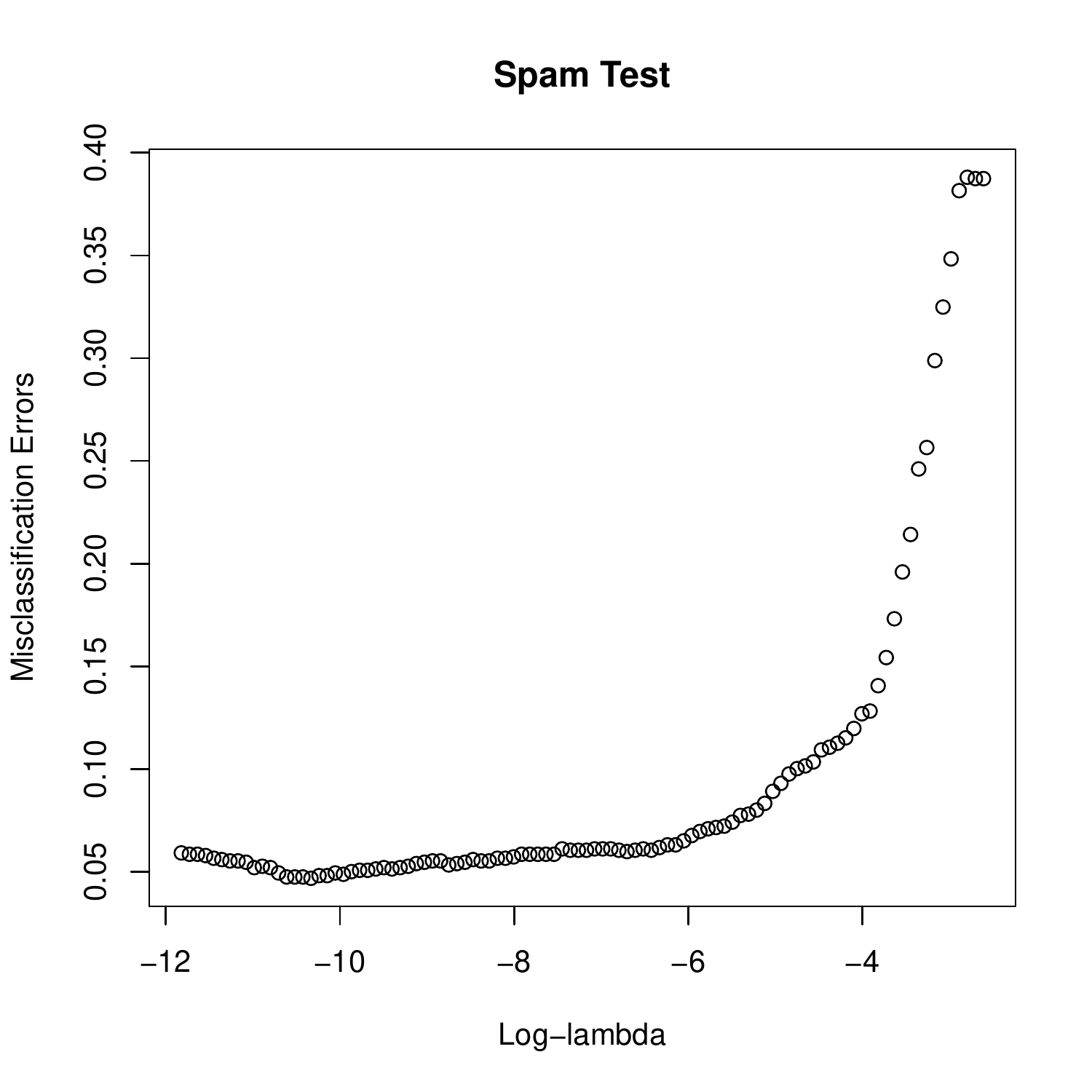}
  \caption{Performance on the spam test data, with 20 knots per
    variable. Increasing to 50 did not make much difference. Minimum
    error is $0.047$.}
\label{fig:spamt}
\end{figure}

The impact is that one could fit a saturated gam model by running say
{\tt glmnet} on the $\{s\}$ bases. An example is given on the spam
data in figure~\ref{fig:spamt}, where the regularization path was
computed at a 100 values of $\lambda$ in seconds.
There are of course some caveats.
\begin{itemize}
\item If you use all the knots for each of $p$ variables, you end up
  with a data matrix of dimension $N\times Np$, which does not scale
  too well. So instead one might use a smaller grid of knots;
  e.g. map the variables onto  $[0,1]$, and then use a grid of say 50 evenly spaced knots
  on this grid, including the end knots.
\item With a large number of knots, the ``variables''  are highly
  correlated, and this can cause numerical issues. The main issue we
  see is that active sets tend to be larger than they should be.
\end{itemize}
Nevertheless, this is an alternative algorithm, which is
closer in spirit to the adaptive splines algorithm.

\section{Proof of Theorem~1}

\addtocounter{theorem}{-1} % huh...
\begin{theorem}{}

Fix $x_1, \ldots, x_n \in [0,1]$ and $f : \reals \rightarrow \reals$ with $f'$ (right-continuous) of bounded total variation, and $f$ constant outside of $[0,1]$.
Then there exists a degree-one saturating spline $\hat{f}$ that matches $f$ on $x_i$ with $\TV(\hat{f}') \le \TV(f').$
\end{theorem}

\begin{proof}
Without loss of generality, we will assume $f(0) = 0$.
Let $\tau = \TV(f')$.
As $f'$ has bounded total variation, there exists a measure $\mu$ on $[0,1]$ such that $f(x) = f_\mu$: \[f(x) =  \int{(x - t)}_+ \;d \mu(t).\]

That is, $f$ is a spline with infinitely many knots. The idea is to use Caratheodory's theorem for convex hulls to see that, as we only care about $\mu$ in terms of its action on a finite number of functions (basically, we only care about the values of $f$ at $x_i$), we can replace $\mu$ with a measure supported on finitely many points.

To make this idea rigorous, note that the vector
\[v = (f(x_1), \ldots, f(x_n), 0) = \int(  (x_1 - t)_+, \ldots, (x_n - t)_+, 1) \;d \mu(t) \]
must lie in convex hull of the (convex) set
\[ C = \{\pm( \tau (x_1 - t)_+, \ldots,  \tau (x_n - t)_+,  \tau) : t \in [0,1] \} \subset \reals^{n+1}\]
as $\| \mu \|_1 = \tau$.
Caratheodory's theorem for convex hulls ensures us that $v$ can be represented as a convex combination of at most $n+2$ points from $C$. Letting these $n+2$ points be represented by their indicies, $t_1, \ldots, t_{n+2}$, and their weights $\alpha_1, \ldots , \alpha_{n+2}$ we define $w_j = \alpha_j \tau$ to obtain:
\begin{align*}
f(x_i)  &= \sum_j w_j (x_i - t_j)_+ = f_\mu(x_i) \\
\sum_j w_j &= 0.
\end{align*}
Here $\mu = \sum_j w_j \delta_{t_j}$.
As $\sum_j | w_j | = \tau$, we have $\TV(f_\mu') = \| \mu \|_1 = \tau$.

\end{proof}
\end{document}